\documentclass[journal]{IEEEtran}

\usepackage[utf8]{inputenc}
\usepackage{color}
\usepackage{xcolor}
\usepackage{array}
\usepackage{verbatim}
\usepackage{float}
\usepackage{amsmath}
\usepackage{amsthm}
\usepackage{amssymb}
\usepackage{graphicx}
\usepackage{longtable}
\usepackage{multirow}
\usepackage{booktabs}
\usepackage[unicode=true,
bookmarks=false,
breaklinks=false,pdfborder={0 0 1},colorlinks=false]
{hyperref}
\hypersetup{
	colorlinks,bookmarksopen,bookmarksnumbered,citecolor=blue,urlcolor=blue}
\usepackage{cite}

\usepackage{lipsum}
\usepackage{mathtools}
\usepackage{cuted}
\providecommand{\tabularnewline}{\\}
\usepackage{algorithmic}
\usepackage{longtable}

\floatstyle{ruled}
\newfloat{algorithm}{tbp}{loa}
\providecommand{\algorithmname}{Algorithm}
\floatname{algorithm}{\protect\algorithmname}

\makeatletter
\let\oldforeign@language\foreign@language
\DeclareRobustCommand{\foreign@language}[1]{%
	\lowercase{\oldforeign@language{#1}}}

\let\oldforeign@language\foreign@language
\DeclareRobustCommand{\foreign@language}[1]{%
	\lowercase{\oldforeign@language{#1}}}

\ifCLASSINFOpdf
\else
\fi

\@ifundefined{showcaptionsetup}{}{%
	\PassOptionsToPackage{caption=false}{subfig}}
\usepackage{subfig}

\usepackage{balance}

\ifCLASSINFOpdf
\else
\fi

\newtheorem{lem}{Lemma}

\newtheorem{thm}{Theorem}
\newtheorem{rem}{Remark}

\newtheorem{assum}{Assumption}
\ifCLASSINFOpdf
\else
\fi

\def\ps@IEEEtitlepagestyle{%
	\def\@oddhead{\parbox[t][\height][t]{\textwidth}{\centering \scriptsize
			Personal use of this material is permitted. Permission from the author(s) and/or copyright holder(s), must be obtained for all other uses. Please contact us and provide details if you believe this document breaches copyrights.\\
			\noindent\makebox[\linewidth]{}
		}\hfil\hbox{}}%
	\def\@evenhead{\scriptsize\thepage \hfil \leftmark\mbox{}}%
	\def\@oddfoot{\parbox[t][\height][l]{\textwidth}{
			\vspace{-20pt}{\rule{\textwidth}{0.4pt}}\\ \footnotesize{\bf{\footnotesize\textcolor{red}{H. Naser, H. A. Hashim, and M. Ahmadi, "Aerial Assistive Payload Transportation Using Quadrotor UAVs with Nonsingular Fast Terminal SMC for Human Physical Interaction," Results in Engineering, vol. 25, pp. 103701, 2025.}}} doi: \href{https://doi.org/10.1016/j.rineng.2024.103701}{10.1016/j.rineng.2024.103701}\\\\
			\noindent\makebox[\linewidth]
		}\hfil\hbox{}}%
	\def\@evenfoot{\MYfooter}}

\makeatother
\begin{document}
	\bstctlcite{IEEEexample:BSTcontrol}

\title{Aerial Assistive Payload Transportation Using Quadrotor UAVs with Nonsingular Fast Terminal SMC for Human Physical Interaction}

\author{Hussein Naser, Hashim A. Hashim, and Mojtaba Ahmadi
	\thanks{This work was supported in part by the National Sciences and Engineering Research Council of Canada (NSERC), under the grants RGPIN-2022-04937.}
	\thanks{H. Naser, H. A. Hashim, and M. Ahmadi are with the Department of Mechanical
		and Aerospace Engineering, Carleton University, Ottawa, Ontario, K1S-5B6,
		Canada (e-mail: hhashim@carleton.ca).}
}



\maketitle
\begin{abstract}
This paper presents a novel approach to utilizing underactuated quadrotor
Unmanned Aerial Vehicles (UAVs) as assistive devices in cooperative
payload transportation task through human guidance and physical interaction.
The proposed system consists of two underactuated UAVs rigidly connected
to the transported payload. This task involves the collaboration between
human and UAVs to transport and manipulate a payload. The goal is
to reduce the workload of the human and enable seamless interaction
between the human operator and the aerial vehicle. An Admittance-Nonsingular
Fast Terminal Sliding Mode Control (NFTSMC) is employed to control
and asymptotically stabilize the system while performing the task,
where forces are applied to the payload by the human operator dictate
the aerial vehicle’s motion. The stability of the proposed controller
is confirmed using Lyapunov analysis. Extensive simulation studies
were conducted using MATLAB, Robot Operating System (ROS), and Gazebo
to validate robustness and effectiveness of the proposed controller
in assisting with payload transportation tasks. Results demonstrates
feasibility and potential benefits utilizing quadrotor UAVs as assistive
devices for payload transportation through intuitive human-guided
control. 
\end{abstract}

\begin{IEEEkeywords}
Cooperative payload transportation, Admittance control, Sliding mode control, Quadrotor control 
\end{IEEEkeywords}

\section{Introduction}\label{sec1}

\subsection{Motivation}
\subsection{Background}
\IEEEPARstart{A}{ssistive} technology has gained significant interest in the field
of robotics in recent years. It plays a crucial role in various domains,
including rehabilitation \cite{ref1,hasan2017cost,naser2023internet,ref2,ref3},
aerial manipulation systems \cite{ref4,ref5}, logistics \cite{ref6,rodriguez2024inspection},
precision agriculture \cite{ref9,meesaragandla2024herbicide}, and
construction \cite{ref10,ref11,ranjbar2023addressing} by enhancing
human capabilities and facilitating complex tasks. Underactuated quadrotor
UAVs with their low cost, simple mechanical design, maneuverability,
and, versatility offer promising potential as assistive devices in
many fields. They can be used for assistive payload transportation
through human physical interaction. This involves collaboration between
human and quadrotors to transport and manipulate a payload. The goal
of such collaborative systems is to reduce the workload of human and
enable intuitive interaction between the operator and the aerial transportation
system. Single quadrotors have been used in payload transportation
to help human in delivering packages and medical supplies \cite{ref12,ref13,ref15}.
Although using a single quadrotor in delivery operations has proven
its effectiveness, it suffers from low weight capacity of the transported
payload \cite{hashim2023exponentially,ref16,hashim2023observer}.
To overcome such problem, researchers have explored cooperative payload
transportation in different configurations such as dual or multiple
agents with suspended-load transportation. Different techniques and
control algorithms have been proposed to achieve the cooperative transportation
task \cite{hashim2023exponentially,ref20,ref21,ref22,shevidi2024quaternion,ref23,ref24,ref25,ref26,hashim2023uwb}.

\subsection{Related Work}

Design of a control algorithm for moving vehicles has been deemed
a challenging task \cite{hashim2023exponentially,fu2023iterative,munoz2019adaptive,hashim2023observer,pisano2011sliding,said2023application,ref17,ref36}.
Tagliabue et al. \cite{ref17} introduced a collaborative object transportation
system using Micro Aerial Vehicles (MAVs) by employing a passive force
control approach. They adopted a master-slave framework, employing
an admittance controller to ensure the slave unit's compliance with
external forces exerted by the master MAV on the payload. To estimate
the external force on the slave, they utilized an Unscented Kalman
Filter depending on data from a Visual-Inertial navigation system.
The system was implemented for outdoor applications to transport a
1.2-meter-long object. Horyna et al. \cite{ref36} introduced a control
approach for a system of two UAVs carrying a beam-type payload for
outdoor applications. A master-slave control architecture comprises
a feedback controller and a Model Predictive Control (MPC) reference
tracker was implemented on the master agent's side, with the slave
agent acting as an actuator following the master's directives. Loianno
and Kumar \cite{ref19} developed a collaborative transportation aerial
system using a team of small quadrotors. They introduced a method
for coordinated control that allows for independent control of each
quadrotor while maintaining the stability of the system. The authors
suggested a cooperative localization strategy that uses the information
on the inherent rigidity of the structure to deduce additional constraints
between the poses of the vehicles. This formulation casts the pose
estimation challenge as an optimization problem on the Lie group of
the Special Euclidean Group $SE(3)$ \cite{hashim2020nonlinear} and
the Lie group of the Special Orthogonal Group $SO(3)$ \cite{hashim2020systematic}.
Rajaeizadeh et al. \cite{ref37} explored the challenge of coordinating
pair of quadrotors rigidly connected to the payload for cooperative
transportation. They introduced a rigid-body formation control system
utilizing the Linear Quadratic Regulator (LQR) method, along with
a Paparazzi-based guidance approach for generating the trajectory
of transportation mission. The system results were validated through
simulations. Mellinger et al. \cite{ref38} examined the challenge
of controlling multiple quadrotors to collaboratively grasp and transport
a payload. The system was modeled using a rigid connection between
multiple UAVs and the payload. The authors conducted an experimental
investigation involving teams of quadrotors working together to grasp,
stabilize, and transport payloads along predefined three-dimensional
paths.

The aforementioned methods of payload transportation depends on preprogrammed
trajectories or remote control, limiting their capabilities in dynamic
environments through physical interaction. Recent advances in control
strategies have paved the way for intuitive human-robot interactions,
allowing seamless collaboration between human operators and autonomous
systems. By integrating admittance control with quadrotors, it becomes
possible to harness the intuitive physical interaction capabilities
of human to guide the motion of the aerial vehicles during payload
transportation tasks. Augugliaro and D’Andrea \cite{ref33} examined
the implementation of admittance control on a single quadrotor for
the physical interaction between a human and an aerial vehicle. Admittance
control enables developers to modify the virtual inertia, damping,
and stiffness of the aerial vehicle facilitating physical interaction.
External forces acting on the quadrotor are measured using sensors
or inferred from position and orientation data and fed into the admittance
controller which adjusts the vehicle's intended path accordingly.
This trajectory is then followed by the inner position and orientation
controllers.

Prajapati and Vashista \cite{ref27} investigated the possible uses
of physical interaction between human and aerial vehicles in outdoor
environments. They introduced a setup involving a rigid object lifted
by both a human and a quadcopter from its extremities. A custom sensor
systems designed to interpret human commands and provide reliable
state feedback. Linearization of the dynamic equations of the system
and state-feedback control techniques were utilized to control the
quadrotor for collaborative payload transportation. Xu et al. \cite{ref29}
introduced a visual impedance control approach for collaborative transportation
between a human and a tethered aerial vehicle. Using the force of
the cable and the features of the visual objects as feedback, the
aerial vehicle follows the human operator. The proposed technique
incorporates a vision-based velocity observer to estimate the relative
velocity between the aerial vehicle and the human. The system is designed
for indoor application where a human participant transported a bar
with the assistance of the tethered aerial vehicle. In Romano et al.
\cite{ref30}, a cooperative payload transport system guided by haptic
feedback was investigated. This system comprises a team of five quadrotors
linked to a payload via tethers. The Human partner can provide haptic
guidance commands by exerting force on the suspended payload,and the
system estimates the applied force and its direction. A centralized
payload controller directs the quadrotors to follow the desired trajectory.

There are several limitations and shortcomings can be identified in
the existing methods of the payload transportation. In one hand, these
methods involve preprogrammed tasks and/or remote-control requirements
which restricts their adaptability and usability in unstructured work
environments. On the other hand, other methods use cables and tethers
in physical interaction configurations, which pose challenges and
difficulties, especially when dealing with aerial vehicles. Cables
can get entangled during missions, causing instability, potential
system damage, and low safety for human partners. Additionally, cable-suspended
payloads suffer from oscillation which introduces complexity to the
system's dynamics and requires complex control algorithms to stabilize
the transportation system. The oscillation also introduces difficulty
to implement the admittance controller for physical interaction, as
it is hard to distinguish between the oscillation and the input of
the interaction forces. Furthermore, this issue significantly affects
battery energy consumption, limiting the mission duration.

\subsection{Motivation and Contribution}

To tackle these challenges, we implemented rigid connections between
the underactuated quadrotor UAVs and the transported payload. This
configuration offers distinct advantages, providing unified dynamics
of the quadrotors and the payload, improving operator safety, and
enabling precise measurement of physical interactions. In this paper,
we propose a novel assistive aerial system in which two quadrotors
are employed for payload transportation following human guidance through
physical interaction. The aerial vehicle lifts the payload to a comfortable
operating altitude, allowing the human operator to interact and guide
the system by applying force directly to the payload without carrying
it. An admittance controller is utilized to translate the forces applied
to the payload by the human operator into motion commands for the
quadrotors, ensuring intuitive and responsive control. The admittance
control algorithm considers the system as a mass-damper-spring system,
where the quadrotors act as dampers and the payload as a mass connected
to a spring. When a force is applied to the payload, the system responds
by moving in the direction of the applied force, simulating the behavior
of a mass-spring-damper system. This approach enables the aerial vehicle
to adaptively adjust its motion based on the forces exerted by the
human operator, providing a natural and intuitive control interface.
The paper contributions are summarized as follows:
\begin{itemize}
	\item Design of an assistive cooperative payload transportation system with
	human physical interaction. 
	\item A model of rigidly connected mechanical system is derived for the
	sake of analysis and control. 
	\item Design and implementation of an admittance controller to enable aerial
	vehicle-human physical interaction. 
	\item Nonsingular Fast Terminal Sliding Mode Control (NFTSMC) is proposed
	to control and asymptotically stabilize the system to track human
	guidance and achieve the transportation task. Lyapunov stability approach
	has been utilized to ensure system stability. 
\end{itemize}

\subsection{Paper Structure}

The rest of the paper is organized as follows: Section \ref{Problem-Formulation}
presents a comprehensive overview of the tools, methods, preliminaries,
and problem formulation. Section \ref{control_section} provides a
detailed explanation of the controller design and stability analysis.
Simulation setup and implementation and numerical results are elaborated
upon in Section \ref{results-discussion}. Finally, Section \ref{Conclusion}
concludes the article and outlines avenues for future work.

\section{Preliminaries and Problem Formulation\label{Problem-Formulation}}

\begin{table*}[t]
	\centering{}\caption{\label{tab:general_symbols} Nomenclature.}
	\begin{tabular}{lll}
		\hline 
		\multicolumn{3}{c}{System related parameters}\tabularnewline
		\hline 
		Symbol & Definition & Unit\tabularnewline
		\hline 
		$\mathcal{I}_{G}$ /$\mathcal{B}_{i}$ & Inertial-frame / $i$ UAV body frame. & -\tabularnewline
		$L_{F}$ & Payload body-fixed-frame. & -\tabularnewline
		$N$ & Number of UAVs. & -\tabularnewline
		$g\in\mathbb{R}$ & Gravitational acceleration. & $[m/s^{2}]$\tabularnewline
		$p_{i}\in\mathbb{R}^{3}$ & $i$ UAV position in $\mathcal{I}_{G}$. & $[m]$\tabularnewline
		$v_{i}\in\mathbb{R}^{3}$ & $i$ UAV linear velocity in $\mathcal{I}_{G}$. & $[m/s]$\tabularnewline
		$\dot{v}_{i}\in\mathbb{R}^{3}$ & $i$ UAV linear acceleration in $\mathcal{I}_{G}$. & $[m/s^{2}]$\tabularnewline
		$m_{i}\in\mathbb{R}$ & $i$ UAV mass. & $[kg]$\tabularnewline
		$J_{i}\in\mathbb{R}^{3\times3}$ & $i$ UAV moment of inertia. & $[kg~m^{2}]$\tabularnewline
		$\Omega_{ji}\in\mathbb{R}$ & Angular speed of $j$ rotor of $i$ quadrotor. & $[rpm]$\tabularnewline
		$k_{t}\in\mathbb{R}$ & Thrust (lift) constant. & $[N/rpm^{2}]$\tabularnewline
		$k_{m}\in\mathbb{R}$ & Moment (drag) constant. & $[N.m/rpm^{2}]$\tabularnewline
		$l\in\mathbb{R}$ & Distance from CoM. & $[m]$\tabularnewline
		$p_{L}$, $v_{L}$, $\dot{v}_{L}$ & Payload position, linear velocity, linear acce. in $\mathcal{I}_{G}$. & \tabularnewline
		$m_{L}$, $J_{L}$ & Payload mass and moment of inertia. & \tabularnewline
		$F_{i}\in\mathbb{R}^{3}$ & $i$ UAV force exerted on payload. & $[N]$\tabularnewline
		$\tau_{i}\in\mathbb{R}^{3}$ & $i$ UAV torque exerted on payload. & $[N.m]$\tabularnewline
		$p_{s}$, $v_{s}$, $\dot{v}_{s}$ & System position, linear velocity, and acce. in $\mathcal{I}_{G}$. & \tabularnewline
		$R\in SO(3)$ & Rotation matrix in body-frame. & -\tabularnewline
		$\omega$, $\dot{\omega}$ & Angular velocity and acce. of the system in $\mathcal{B}_{i}$. & -\tabularnewline
		$(\phi,\theta,\psi)$ & Roll, pitch, and yaw Euler angles. & $[rad]$\tabularnewline
		$m_{t}$, $J_{t}$ & Total mass and moment of inertia of entire system. & -\tabularnewline
		$F_{t}\in\mathbb{R}$ & Total thrust produced by two quadrotors. & $[N]$\tabularnewline
		$U_{t}\in\mathbb{R}^{3}$ & Total (rolling, pitching, yawing) moments. & $[N.m]$\tabularnewline
		$F_{drag}\in\mathbb{R}^{3}$ & Aerodynamic drag force acting on the system. & $[N]$\tabularnewline
		$M_{drag}\in\mathbb{R}^{3}$ & Aerodynamic drag moment acting on the system. & $[N.m]$\tabularnewline
		$D_{l}\in\mathbb{R}^{3}$ & Translational external disturbances. & $[N]$\tabularnewline
		$D_{r}\in\mathbb{R}^{3}$ & Rotational external disturbances. & $[N.m]$\tabularnewline
		\hline 
		\hline 
		\multicolumn{3}{c}{Controller Related Parameters and Symbols}\tabularnewline
		\hline 
		Symbol & Definition & Unit\tabularnewline
		\hline 
		$\chi\in\mathbb{R}^{3}$ & Actual position and orientation of the entire system & -\tabularnewline
		$\chi_{d}\in\mathbb{R}^{3}$ & Desired position and orientation of the entire system & -\tabularnewline
		$e\in\mathbb{R}^{6}$ & Tracking error & -\tabularnewline
		$S\in\mathbb{R}^{6}$ & Sliding surface & -\tabularnewline
		$\xi\in\mathbb{R}^{6\times6}$ & Positive diagonal matrix of control constants & -\tabularnewline
		$\eta\in\mathbb{R}^{6\times6}$ & Positive diagonal matrix of control constants & -\tabularnewline
		$a\in\mathbb{R}$ & Positive constant & -\tabularnewline
		$\lambda_{1}\in\mathbb{R}^{6\times6}$ & Positive diagonal matrix of control constants & -\tabularnewline
		$\lambda_{2}\in\mathbb{R}^{6\times6}$ & Positive diagonal matrix of control constants & -\tabularnewline
		$V>0$ & Positive definite Lyapunov function for the system & -\tabularnewline
		$(u_{x},u_{y},u_{z})$ & Virtual control inputs & -\tabularnewline
		$U_{1}\in\mathbb{R}$ & Designed total thrust control input & $[N]$\tabularnewline
		$U_{2}\in\mathbb{R}$ & Designed total rolling moment of entire system & $[N.m]$\tabularnewline
		$U_{3}\in\mathbb{R}$ & Designed total pitching moment of entire system & $[N.m]$\tabularnewline
		$U_{4}\in\mathbb{R}$ & Designed total yawing moment of entire system & $[N.m]$\tabularnewline
		$\mathbb{M}\in\mathbb{R}^{3\times3}$ & Virtual mass for admittance controller & $[kg]$\tabularnewline
		$\mathbb{C}\in\mathbb{R}^{3\times3}$ & Virtual damping coefficients for admittance controller & $[N.s/m]$\tabularnewline
		$\mathbb{K}\in\mathbb{R}^{3\times3}$ & Virtual spring constants for admittance controller & $[N.s/m]$\tabularnewline
		$\mathbb{T}_{d}\in\mathbb{R}^{3}$ & Desired trajectory of admittance controller & $[m]$\tabularnewline
		$\mathbb{T}_{r}\in\mathbb{R}^{3}$ & Reference trajectory generated by admittance controller & $[m]$\tabularnewline
		\hline 
	\end{tabular}
\end{table*}

In this work, our objective is to design and implement an assistive
aerial system to assist humans in payload transportation with human
physical interaction and guidance. The proposed system consists of
two quadrotors collaboratively lifting and transporting a common payload
that is rigidly connected to the quadrotors. The human operator can
interact with the assistive aerial system through force feedback,
allowing intuitive control of the system during payload transportation
tasks. The human operator can directly push or apply forces to the
transported payload to guide the system in any direction in three-dimensional
(3D) space. The applied force can be measured using either a one force-torque
sensor attached to the center of mass (CoM) of the payload or two
force-torque sensors attached at the points of the rigid connection
of each quadrotor with the payload ends. The dynamic model of each
component of the assistive system (two quadrotors and the payload)
will be given separately; then the dynamic model for the entire system
will be derived in detail for control purposes. Without loss of generality,
the entire system will be considered as one rigid body and treated
as one aerial vehicle. All the notations, parameters, and symbols
used in this paper are illustrated in the following tables. For notations
and symbols related to the modeling of individual quadrotor, payload,
entire assistive system, controller-related notations and symbols
respectively, refer to Table \ref{tab:general_symbols}.

\subsection{System Model}

For the modeling purposes, we assign a global world inertial reference
frame $\mathcal{I}_{G}=\{O,X,Y,Z\}$ and a body-fixed reference frame
$L_{F}=\{o,x_{L},y_{L},z_{L}\}$ attached to the CoM of the payload
as well as a reference frame $\mathcal{B}_{i}=\{o_{i},x_{B_{i}},y_{B_{i}},z_{B_{i}}\}$
attached to the CoM of each quadrotor where $i=1....N$ ($N=2$ for
the system in this work). All reference frames have a positive $z$
axis pointing upward (gravity is negative). Let $p_{i}=[x_{i},y_{i},z_{i}]^{\top}$
and $v_{i}=\dot{p}$ be the translational position and velocity of
the quadrotor, respectively, defined in $\mathcal{I}_{G}$, $\omega=[p,q,r]^{\top}$
be the vector of angular rates defined in the body-fixed frame, and
$\Theta=[\phi,\theta,\psi]^{\top}$ be the vector of Euler angles,
roll, pitch, and yaw that represent the attitude of the aerial vehicle
in $\mathcal{I}_{G}$. $R\in SO(3)$ represents the rotation matrix
from the body-fixed frame to $\mathcal{I}_{G}$ described with respect
to the Special Orthogonal Group $SO(3)$ where $R$ has a determinant
of $+1$ and $RR^{\top}=\mathbf{I}_{3}$ with $\mathbf{I}_{3}$ being
an identity matrix with dimension $3$-by-$3$ \cite{hashim2019special,hashim2020systematic,hashim2019nItoStrat}.
The rotation matrix $R$ follows the convention of $Z-X-Y$ of Euler
angles and is given by \cite{hashim2019special}: 
\begin{equation}
	\mathbf{R}=\begin{bmatrix}c\theta c\psi-s\phi s\theta s\psi & -c\phi s\psi & s\theta c\psi+s\phi c\theta s\psi\\
		c\theta s\psi+s\phi s\theta c\psi & c\phi c\psi & s\theta s\psi-s\phi c\theta c\psi\\
		-c\phi s\theta & s\phi & c\phi c\theta
	\end{bmatrix}\label{R_equ}
\end{equation}
where $c\phi$ and $s\phi$ denote $cos(\phi)$ and $sin(\phi)$,
respectively. The relationship between $\omega$ and the rate of change
of the Euler angles $\dot{\Theta}=[\dot{\phi},\dot{\theta},\dot{\psi}]^{\top}$
is described by \cite{hashim2019special}: 
\begin{equation}
	\begin{bmatrix}p\\
		q\\
		r
	\end{bmatrix}=\begin{bmatrix}c\theta & 0 & -c\phi s\theta\\
		0 & 1 & s\phi\\
		s\theta & 0 & c\phi c\theta
	\end{bmatrix}\begin{bmatrix}\dot{\phi}\\
		\dot{\theta}\\
		\dot{\psi}
	\end{bmatrix}\label{omegat_theta_equ}
\end{equation}

\subsubsection{Individual Quadrotor Dynamics}

\begin{figure}
	\centering{}\includegraphics[clip,width=0.5\textwidth]{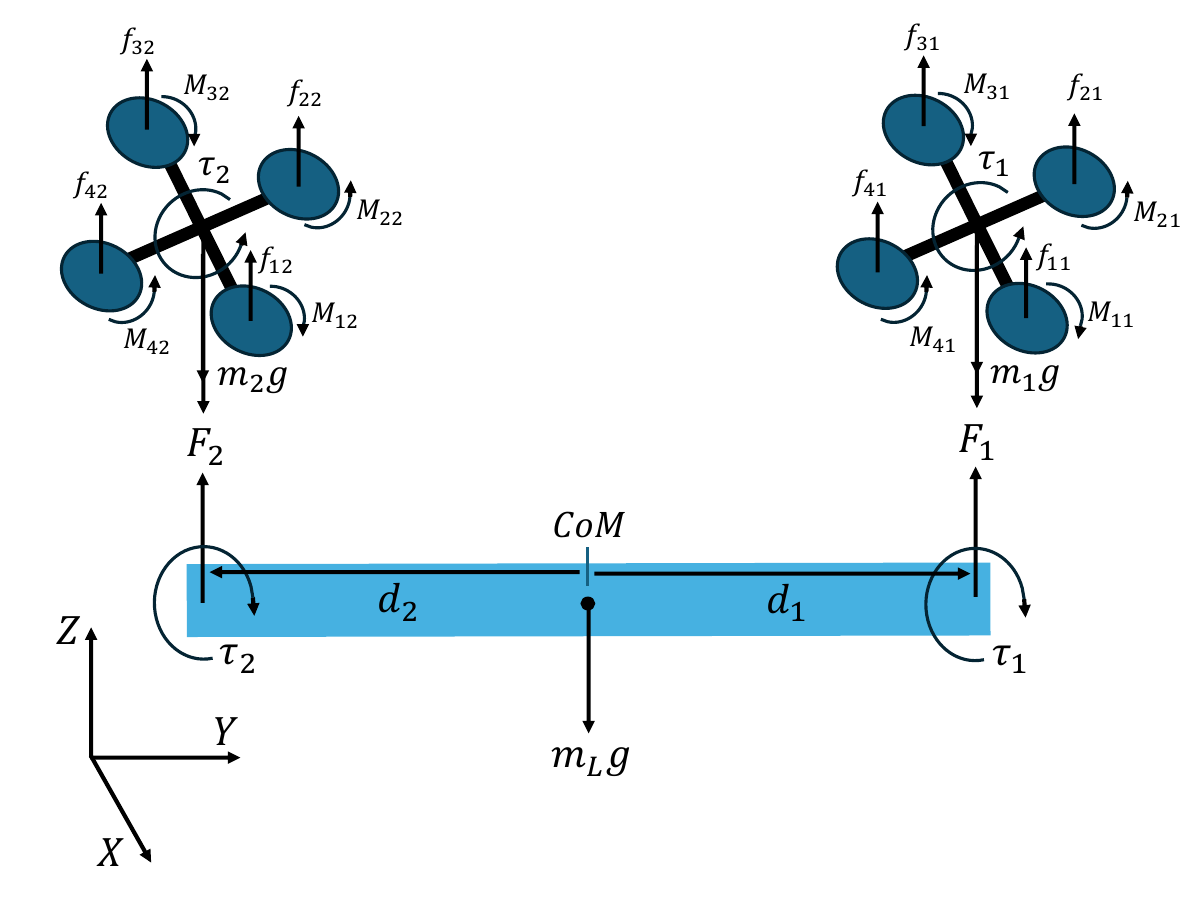}
	\caption{\label{fig1}Free body diagram of the entire system components.}
\end{figure}

Using the Newton-Euler method and referring to Figure \ref{fig1},
the translational and rotational dynamic model of the individual underactuated
quadrotor UAV is given as follows \cite{hashim2023exponentially,hashim2023observer}:
\begin{equation}
	\begin{aligned}m_{i}\dot{v}_{i} & =R\begin{bmatrix}0\\
			0\\
			u_{1_{i}}
		\end{bmatrix}-\begin{bmatrix}0\\
			0\\
			m_{i}g
		\end{bmatrix}-F_{i}\\
		J_{i}\dot{\omega} & =\begin{bmatrix}u_{2_{i}}\\
			u_{3_{i}}\\
			u_{4_{i}}
		\end{bmatrix}-\begin{bmatrix}p\\
			q\\
			r
		\end{bmatrix}\times J_{i}\begin{bmatrix}p\\
			q\\
			r
		\end{bmatrix}-\tau_{i}
	\end{aligned}
	\label{trans_rot_quad_equ}
\end{equation}
where $m_{i}\in\mathbb{R}$ is the mass, $J_{i}\in\mathbb{R}^{3\times3}$
is the moment of inertia, $\dot{v_{i}}\in\mathbb{R}^{3}$ is the translational
acceleration, $u_{1_{i}}\in\mathbb{R}$ is the total thrust measured
in the body-fixed reference frame, and $[u_{2_{i}},u_{3_{i}},u_{4_{i}}]^{\top}$
are the rolling, pitching, and yawing moments in $x_{B_{i}},y_{B_{i}},z_{B_{i}}$
axes of each individual quadrotor, respectively, $g\in\mathbb{R}$
is the gravity, $\dot{\omega}\in\mathbb{R}^{3}$ is the angular acceleration
which is the same for all components of the system due to rigid connection,
and $F_{i}\in\mathbb{R}^{3}$ and $\tau_{i}\in\mathbb{R}^{3}$ are
the force and torque of the rigid interconnection between the two
quadrotors and the payload, respectively. 

The total thrust and moments of individual quadrotor are given as
follows: 
\begin{equation}
	\begin{aligned}u_{1_{i}} & =\sum_{j=1}^{4}f_{ji}\\
		\begin{bmatrix}u_{2_{i}}\\
			u_{3_{i}}\\
			u_{4_{i}}
		\end{bmatrix} & =\begin{bmatrix}l(f_{2_{i}}-f_{4_{i}})\\
			l(f_{3_{i}}-f_{1_{i}})\\
			M_{1_{i}}-M_{2_{i}}+M_{3_{i}}-M_{4_{i}}
		\end{bmatrix}
	\end{aligned}
	\label{total_i_thrust_moments_equ}
\end{equation}
where $(f_{ji}=k_{t}\Omega_{ji}^{2})$ is the thrust generated by
each rotor of individual quadrotor, and it is proportional to the
square of the angular speed $(\Omega_{ji})$ of the rotor, $k_{t}$
is the thrust (lift) constant, $(M_{ji}=k_{m}\Omega_{ji}^{2})$ is
the moment produced by each rotor, $k_{m}$ is the moment (drag) constant,
and $l$ is the distance from the CoM of the quadrotor to the axis
of rotation of the rotors. The total thrust and moments in Equation
(\ref{total_i_thrust_moments_equ}) can be combined to form the input
vector for each individual quadrotor as given in Equation (\ref{input_vector_equ})
\cite{hashim2023exponentially,shevidi2024quaternion}: 
\begin{equation}
	\begin{bmatrix}u_{1_{i}}\\
		u_{2_{i}}\\
		u_{3_{i}}\\
		u_{4_{i}}
	\end{bmatrix}\begin{bmatrix}1 & 1 & 1 & 1\\
		0 & l & 0 & -l\\
		-l & 0 & l & 0\\
		\mu & -\mu & \mu & -\mu
	\end{bmatrix}\begin{bmatrix}f_{1_{i}}\\
		f_{2_{i}}\\
		f_{3_{i}}\\
		f_{4_{i}}
	\end{bmatrix}\label{input_vector_equ}
\end{equation}
where $(\mu=k_{m}/k_{t})$ is the relationship between the thrust
and moment constants.

\subsubsection{Payload Dynamics}

Let $p_{L}=[x_{L},y_{L},z_{L}]^{\top}$ and $v_{L}=\dot{p}_{L}$ be
the translational position and velocity of the CoM of the payload,
respectively, defined in the inertial reference frame. Referring to
Figure \ref{fig1}, the translational and rotational dynamics of the
payload are given as follows: 
\begin{equation}
	\begin{aligned}m_{L}\dot{v}_{L} & =(F_{1}+F_{2})-\begin{bmatrix}0\\
			0\\
			m_{L}g
		\end{bmatrix}\\
		J_{L}\dot{\omega} & =(\tau_{1}+\tau_{2})-\begin{bmatrix}p\\
			q\\
			r
		\end{bmatrix}\times J_{L}\begin{bmatrix}p\\
			q\\
			r
		\end{bmatrix}+(d_{1}\times F_{1})+(d_{2}\times F_{2})
	\end{aligned}
	\label{trans_rot_payload_equ}
\end{equation}
where $m_{L}\in\mathbb{R}$ is the mass, $J_{L}\in\mathbb{R}^{3\times3}$
is the diagonal matrix of the moment of inertia, $\dot{v}_{L}\in\mathbb{R}^{3}$
is the translational acceleration, and $(F_{1}$ and $F_{2})\in\mathbb{R}^{3}$
and $(\tau_{1}$ and $\tau_{2})\in\mathbb{R}^{3}$ are the forces
and torques exerted by the two quadrotors on the payload at the points
of contact of the rigid connection. The expressions $(d_{1}\times F_{1})$
and $(d_{2}\times F_{2})$ express the moments due to the forces $(F_{1}$
and $F_{2})$ at the points of contact and $d_{1}$ and $d_{2}$ are
the vectors from the CoM of the payload to the CoM of quadrotor1 and
quadrotor2, respectively, where $(d_{i}=p_{i}-p_{L})$.

\subsubsection{System Kinematics}

The kinematic relationships between the payload and the two quadrotors
are given in Equation (\ref{kinematic_equ}): 
\begin{equation}
	\begin{cases}
		p_{i} & =p_{L}+Rd_{i},\\
		\dot{p}_{i} & =\dot{p}_{L}+\omega\times d_{i},\\
		\ddot{p}_{i} & =\ddot{p}_{L}+\dot{\omega}\times d_{i}+\omega(\omega\times d_{i}),
	\end{cases}\label{kinematic_equ}
\end{equation}

\subsubsection{Dynamics of the entire system}

To formulate the dynamic equations of the entire system, let Assumption
\ref{as:1} hold true:

\begin{assum}\label{as:1} In this work, we assume the following. 
	\begin{itemize}
		\item The system is composed of three separate rigid components that are
		rigidly connected with their own specific physical characteristics. 
		\item The two quadrotors are identical with the same known geometrical and
		physical specifications. 
		\item The payload is a circular section beam with a mass of $m_{L}$, a
		length of $L$, and a cross-sectional radius $r_{L}$. 
		\item The entire system is symmetric in $X$ and $Y$ axes. 
	\end{itemize}
\end{assum}

Let $p_{s}=[x,y,z]^{\top}$ and $v_{s}=\dot{p}_{s}$ be the translational
position and velocity of the CoM of the entire aerial vehicle, respectively,
defined in the inertial reference frame. Based on Assumption \ref{as:1}
and referring to Figure \ref{fig1}, combining the dynamics of the
quadrotor in Equation (\ref{trans_rot_quad_equ}) with the dynamics
of the payload in Equation (\ref{trans_rot_payload_equ}) as well
as adding the aerodynamic drag effects and external disturbances results
in formulation of the dynamics of the entire system as follows: 
\begin{equation}
	\begin{aligned}m_{t}\dot{v}_{s} & =R\begin{bmatrix}0\\
			0\\
			F_{t}
		\end{bmatrix}-\begin{bmatrix}0\\
			0\\
			m_{t}g
		\end{bmatrix}-F_{drag}+D_{l}\\
		J_{t}\dot{\omega} & =U_{t}-\begin{bmatrix}p\\
			q\\
			r
		\end{bmatrix}\times J_{t}\begin{bmatrix}p\\
			q\\
			r
		\end{bmatrix}-M_{drag}+D_{r}
	\end{aligned}
	\label{trans_rot_sys_equ}
\end{equation}
where $m_{t}\in\mathbb{R}$ is the total mass of the entire system,
and it can be given as $(m_{t}=m_{1}+m_{2}+m_{L})$, $\dot{v}_{s}\in\mathbb{R}^{3}$
is the translational acceleration of the entire system, $J_{t}\in\mathbb{R}^{3\times3}$
is the total moment of inertia of the entire system, $D_{l}\in\mathbb{R}^{3}$
and $D_{r}\in\mathbb{R}^{3}$ are translational and rotational external
disturbances, $(F_{drag}=K_{l_{drag}}v)\in\mathbb{R}^{3}$ and $(M_{drag}=K_{r_{drag}}\omega)\in\mathbb{R}^{3}$
are the aerodynamic drag forces and torques, $K_{l_{drag}}$ and $K_{r_{drag}}$
are positive diagonal matrices that represent the translational and
rotational drag coefficients of the system, respectively, $F_{t}\in\mathbb{R}$
and $U_{t}\in\mathbb{R}^{3}$ are the total thrust and moments produced
by the two quadrotors measured in the body-fixed frame of the system.
Considering each quadrotor produces thrust and moments in its own
coordinate frame, a relationship between the behavior of the entire
system and the quadrotors needs to be developed depending on the system's
configuration as follows: 
\begin{equation}
	\begin{bmatrix}F_{t}\\
		U_{t}
	\end{bmatrix}=\mathbf{B}u_{q}\label{contro_input_equ}
\end{equation}
where $\mathbf{B}\in\mathbb{R}^{4\times4N}$ is a constant matrix
and determined based on the system configuration and the yaw angle
$(\psi_{i})$ of each quadrotor with respect to the payload which
is designed to be zero $(\psi_{i}=0)$ for the rigid connection of
the system. $\mathbf{B}$ is given as follows: 
\begin{equation}
	\mathbf{B}=\sum_{i=1}^{2}\begin{bmatrix}1 & 0 & 0 & 0\\
		d_{i}(2) & 1 & 0 & 0\\
		-d_{i}(1) & 0 & 1 & 0\\
		0 & 0 & 0 & 1
	\end{bmatrix}\label{b_matrix_equ}
\end{equation}
and $u_{q}\in\mathbb{R}^{4N}$ is the vector of the control input
of the two quadrotors as given below: 
\begin{equation}
	u_{q}=[u_{11},u_{21},u_{31},u_{41},u_{12},u_{22},u_{32},u_{42}]^{\top}\label{u_inputs}
\end{equation}

\begin{rem}
	\label{rem_1}In our control scheme, the $[F_{t},U_{t}]^{\top}$ is
	firstly calculated then used to find $(u_{q})$ in (\ref{u_inputs})
	for the quadrotors to follow the desired tracking control. It can
	be clearly noticed that the system in Equation (\ref{contro_input_equ})
	is underdetermined, where there are only four equations for eight
	unknowns. Therefore, we need to optimize the solutions through minimizing
	a cost function $\mathcal{L}(u_{q}):\mathbb{R}^{8}\rightarrow\mathbb{R}$
	such that:
	\begin{equation}
		u_{q}^{*}=\text{argmin}\{\mathcal{L}|[F_{t},~U_{t}]^{\top}=\mathbf{B}u_{q}\}\label{optim_fun_equ}
	\end{equation}
	where the cost function $\mathcal{L}$ in (\ref{optim_fun_equ}) is
	given as follows:
	\begin{equation}
		\mathcal{L}=\sum_{i=1}^{2}c_{1_{i}}u_{1_{i}}+c_{2_{i}}u_{2_{i}}+c_{3_{i}}u_{3_{i}}+c_{4_{i}}u_{4_{i}}\label{cost_fun_equ}
	\end{equation}
	where $c_{ji}$ are the cost function's coefficients that can be used
	to define a matrix $\mathcal{H}\in\mathbb{R}^{8\times8}$ such that
	the cost function $\mathcal{L}$ in (\ref{cost_fun_equ}) can be redefined
	as $(\mathcal{L}=\|\mathcal{H}u_{q}\|_{2}^{2})$, and $\mathcal{H}$
	is given as follows:
	\begin{equation}
		\mathcal{H}=\sqrt{diag(c_{11},c_{21},c_{31},c_{41},c_{12},c_{22},c_{32},c_{42})}\label{h_matrix_equ}
	\end{equation}
	Using $\mathbf{B}$ in (\ref{b_matrix_equ}) and $\mathcal{H}$ in
	(\ref{h_matrix_equ}), the solution can be determined using pseudo
	inverse (Moore-Penrose inverse) as \cite{ref38}: 
	\begin{equation}
		\begin{split}u_{q}^{*} & =\mathcal{H}^{-1}(\mathbf{B}\mathcal{H}^{-1})^{+}[F_{t},~U_{t}]^{\top}\\
			& =\mathcal{H}^{-2}\mathbf{B}^{\top}(\mathbf{B}\mathcal{H}^{-2}\mathbf{B}^{\top})^{-1}[F_{t},~U_{t}]^{\top}
		\end{split}
		\label{optim_cont_final_equ}
	\end{equation}
	where $+$ denotes the pseudo inverse and $u_{q}^{*}$ in (\ref{optim_cont_final_equ})
	satisfies Remark \ref{rem_1}.
\end{rem}

\section{Controller Design and Stability Analysis\label{control_section}}

\begin{figure*}[h!]
	\centering{}\includegraphics[width=0.85\textwidth]{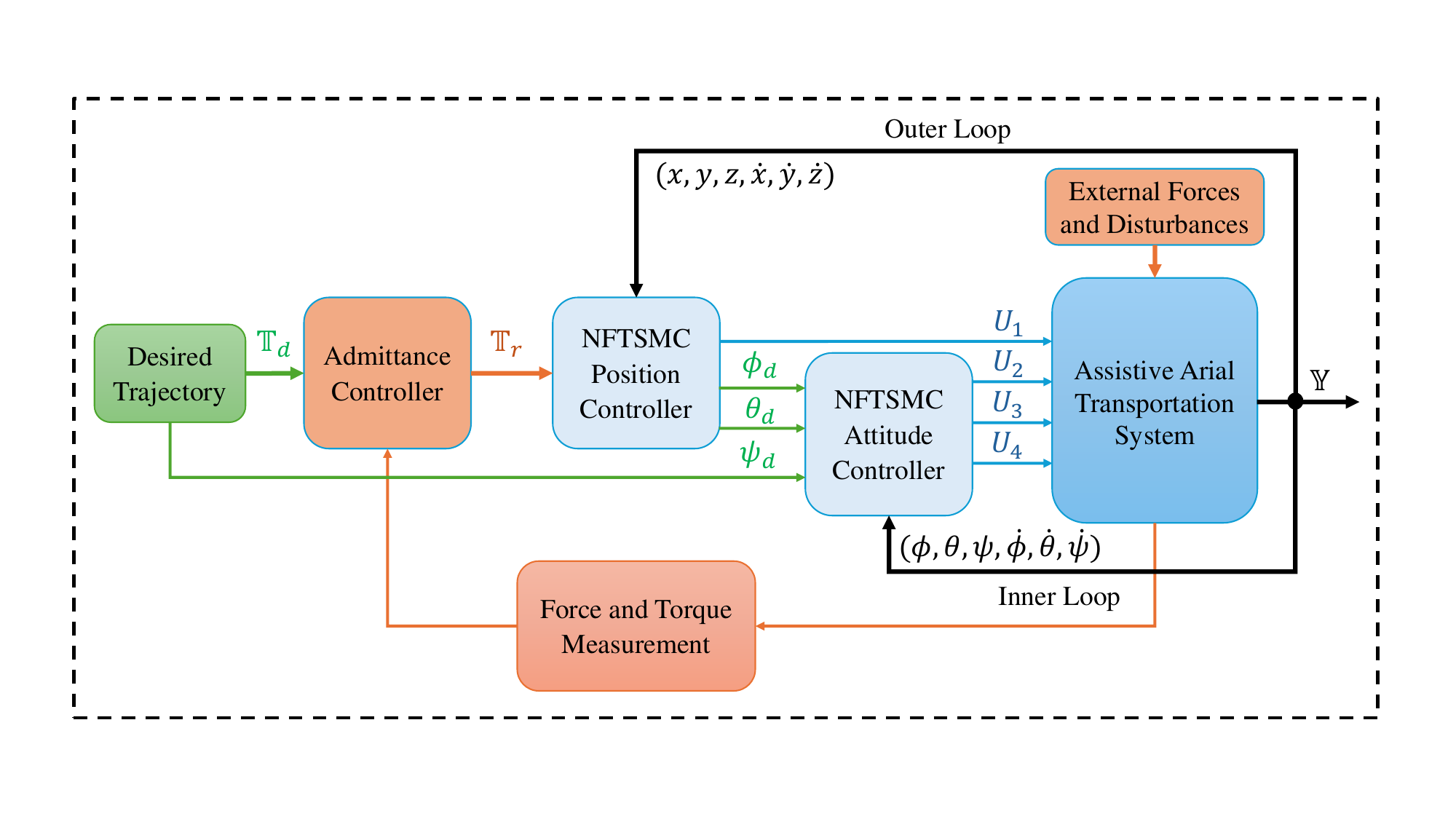}
	\caption{\label{fig2}Schematic block diagram of the control system.}
\end{figure*}

In this section, the controller will be designed to control the assistive
payload transportation system with human physical interaction to track
human guidance and stabilize the system while transporting the payload
in the presence of aerodynamic drag forces and external disturbances.
Figure \ref{fig2} depicts a schematic diagram of the overall control
system. The controller will be designed based on a novel combination
of the admittance controller for human-aerial vehicle physical interaction
and NFTSMC, which is a well-known nonlinear controller in terms of
its robustness against modeling uncertainties and external disturbances,
as well as fast convergence to ensure system stability \cite{hashim2023exponentially,pisano2011sliding,shevidi2024quaternion}.

For control design purposes, the dynamic model in Equation (\ref{trans_rot_sys_equ})
will be expanded and written in the following form: 
\begin{equation}
	\ddot{\chi}=\digamma+\delta+bu\label{xsi_equ}
\end{equation}
where $\ddot{\chi}=[\ddot{x},\ddot{y},\ddot{z},\ddot{\phi},\ddot{\theta},\ddot{\psi}]^{\top}$
is the vector of the system translational and angular accelerations,
$\digamma$ and $b\neq0$ are smooth non-linear functions, and $\delta$
represents the uncertainties and external forces and disturbances,
and $u$ is the control input.

\begin{assum} \label{as:2} $\delta$ in Equation (\ref{xsi_equ})
	is assumed to be bounded and satisfy $\|\delta\|\leqslant\varpi$,
	where $\varpi>0$. \end{assum}

\begin{assum} \label{as:3} The desired position of the entire system,
	$X_{d}=[x_{d},y_{d},z_{d}]$, is bounded, smooth, and differentiable.
\end{assum}

\begin{assum} \label{as:4} For small $(\phi,\theta,\psi)$ angles
	applications such as the system in this work, the Euler angles rate,
	$\dot{\Theta}=(\dot{\phi},\dot{\theta},\dot{\psi})$ is assumed to
	be equal to the angular velocity $\omega=(p,q,r)$. \end{assum}

Let Assumptions \ref{as:2}, \ref{as:3}, and \ref{as:4} hold true,
the extended dynamic equations of the system are given as follows:
\begin{equation}
	\begin{cases}
		\ddot{x} & =\frac{1}{m_{t}}[-k_{l_{drag_{x}}}\dot{x}+d_{l_{1}}+(s\theta c\psi+s\phi c\theta s\psi)U_{1}]\\
		\ddot{y} & =\frac{1}{m_{t}}[-k_{l_{drag_{y}}}\dot{y}+d_{l_{2}}+(s\theta s\psi-s\phi c\theta c\psi)U_{1}]\\
		\ddot{z} & =\frac{1}{m_{t}}[-k_{l_{drag_{z}}}\dot{z}+d_{l_{3}}-m_{t}g+(c\phi c\theta)U_{1}]\\
		\ddot{\phi} & =\frac{1}{J_{x}}[(J_{y}-J_{z})\dot{\theta}\dot{\psi}-k_{r_{{drag}_{\phi}}}\dot{\phi}+d_{r_{1}}+U_{2}]\\
		\ddot{\theta} & =\frac{1}{J_{y}}[(J_{z}-J_{x})\dot{\phi}\dot{\psi}-k_{r_{{drag}_{\theta}}}\dot{\theta}+d_{r_{2}}+U_{3}]\\
		\ddot{\psi} & =\frac{1}{J_{z}}[(J_{x}-J_{y})\dot{\phi}\dot{\theta}-k_{r_{{drag}_{\psi}}}\dot{\psi}+d_{r_{3}}+U_{4}]
	\end{cases}\label{xsi_seprate_equ}
\end{equation}

\begin{rem}
	\label{remark_subsystem}As it can be noticed from Equation (\ref{xsi_seprate_equ})
	that the system is divided into two subsystems, namely, the position
	subsystem (first three lines in Equation (\ref{xsi_seprate_equ}))
	is underactuated subsystem with one control input $(U_{1})$ and three
	outputs $(x,y,z)$, and the orientation subsystem (last three lines
	in Equation (\ref{xsi_seprate_equ})) is full actuated subsystem with
	three control inputs $(U_{2},U_{3},U_{4})$ and three outputs $(\phi,\theta,\psi)$.
	Therefore, the controller design needs to account for such challenging
	dynamical system. 
\end{rem}
In order to deal with the challenge of the underactuated subsystem
in Equation (\ref{xsi_seprate_equ}) as remarked in Remark \ref{remark_subsystem},
let us consider the following virtual control inputs: 
\begin{equation}
	\begin{bmatrix}u_{x}\\
		u_{y}\\
		u_{z}
	\end{bmatrix}=\begin{bmatrix}(s\theta c\psi+s\phi c\theta s\psi)U_{1}\\
		(s\theta s\psi-s\phi c\theta c\psi)U_{1}\\
		(c\phi c\theta)U_{1}
	\end{bmatrix}\label{vertual_cont_equ}
\end{equation}
Therefore, components of Equation (\ref{xsi_equ}) are given as follows:
\begin{equation}
	\begin{split}\digamma= & \begin{bmatrix}-\frac{1}{m_{t}}k_{l_{drag_{x}}}\dot{x}\\
			-\frac{1}{m_{t}}k_{l_{drag_{y}}}\dot{y}\\
			-g-\frac{1}{m_{t}}k_{l_{drag_{z}}}\dot{z}\\
			\frac{(J_{y}-J_{z})}{J_{x}}\dot{\theta}\dot{\psi}-\frac{k_{r_{drag_{\phi}}}}{J_{x}}\dot{\phi}\\
			\frac{(J_{z}-J_{x})}{J_{y}}\dot{\phi}\dot{\psi}-\frac{k_{r_{drag_{\theta}}}}{J_{y}}\dot{\theta}\\
			\frac{(J_{x}-J_{y})}{J_{z}}\dot{\phi}\dot{\theta}-\frac{k_{r_{drag_{\psi}}}}{J_{z}}\dot{\psi}
		\end{bmatrix}\\
		\delta= & \begin{bmatrix}\frac{d_{l_{1}}}{m_{t}},\frac{d_{l_{2}}}{m_{t}},\frac{d_{l_{3}}}{m_{t}},\frac{d_{r_{1}}}{J_{x}},\frac{d_{r_{2}}}{J_{y}},\frac{d_{r_{3}}}{J_{z}}\end{bmatrix}^{\top}\\
		b= & \begin{bmatrix}\frac{1}{m_{t}},\frac{1}{m_{t}},\frac{1}{m_{t}},\frac{1}{J_{x}},\frac{1}{J_{y}},\frac{1}{J_{z}}\end{bmatrix}^{\top}
	\end{split}
	\label{components_equ}
\end{equation}

\subsection{Sliding surface design}

Let $e\in\mathbb{R}^{6}$ be the tracking error of the system which
is given as: 
\begin{equation}
	e=\chi_{d}-\chi\label{error_equ}
\end{equation}
where $\chi=[x,y,z,\phi,\theta,\psi]^{\top}$and $\chi_{d}=[x_{d},y_{d},z_{d},\phi_{d},\theta_{d},\psi_{d}]^{\top}$
are the actual and desired position and orientation of the entire
system, respectively. For the system in Equation (\ref{xsi_equ})
and the tracking error in Equation (\ref{error_equ}), the Non-singular
Fast Terminal Sliding Mode (NFTSM) surfaces are considered as follows:
\begin{equation}
	\begin{cases}
		S_{x} & =\dot{e_{x}}+\xi e_{x}+\eta sgn(e_{x})^{a}\\
		S_{y} & =\dot{e_{y}}+\xi e_{y}+\eta sgn(e_{y})^{a}\\
		S_{z} & =\dot{e_{z}}+\xi e_{z}+\eta sgn(e_{z})^{a}\\
		S_{\phi} & =\dot{e_{\phi}}+\xi e_{\phi}+\eta sgn(e_{\phi})^{a}\\
		S_{\theta} & =\dot{e_{\theta}}+\xi e_{\theta}+\eta sgn(e_{\theta})^{a}\\
		S_{\psi} & =\dot{e_{\psi}}+\xi e_{\psi}+\eta sgn(e_{\psi})^{a}
	\end{cases}
\end{equation}
where $\xi>0$, ~$\eta>0$, ~$a\geq1$, $sgn(e)^{a}=|e|^{a}sgn(e)$
and its derivative is given as \cite{ref28} $\frac{d}{dt}(sgn(e)^{a})=a|e|^{a-1}\dot{e}$,
and $sgn(.)$ is the sign function which is given in Equation (\ref{sgn_equ}):
\begin{equation}
	sgn(e):=\begin{cases}
		1 & \text{if }e>0,\\
		0 & \text{if }e=0,\\
		-1 & \text{if }e<0
	\end{cases}\label{sgn_equ}
\end{equation}
For simplicity of analysis and calculation, let $S=[S_{x},S_{y},S_{z},S_{\phi},S_{\theta},S_{\psi}]^{\top}$
represent the general NFTSM surface of all system variables above,
and it is given as follows: 
\begin{equation}
	S=\dot{e}+\xi e+\eta|e|^{a}sgn(e)\label{sliding_surface_equ}
\end{equation}
Taking the time derivative of the sliding surface in Equation (\ref{sliding_surface_equ})
results in: 
\begin{equation}
	\begin{split}\dot{S} & =\ddot{e}+\xi\dot{e}+\eta a|e|^{a-1}\dot{e}\\
		& =\ddot{\chi_{d}}-\ddot{\chi}+\xi\dot{e}+\eta a|e|^{a-1}\dot{e}
	\end{split}
	\label{sliding_surface_dot_equ}
\end{equation}
Substituting Equation (\ref{xsi_equ}) into Equation (\ref{sliding_surface_dot_equ})
results in: 
\begin{equation}
	\dot{S}=\ddot{\chi_{d}}-\digamma-\delta-bu+(\xi+\eta a|e|^{a-1})\dot{e}\label{sdot_equ}
\end{equation}
Considering $\hat{u}$ as an equivalent controller (continuous control
law) to achieve $\dot{S}=0$, results in the following: 
\begin{equation}
	\hat{u}=\frac{1}{b}\begin{bmatrix}\ddot{\chi_{d}}-\digamma-\delta+(\xi+\eta a|e|^{a-1})\dot{e}\end{bmatrix}\label{u_hat_equ}
\end{equation}
To address dynamic uncertainties and external disturbances, as well
as to increase the speed of convergence to the sliding surface, the
control law can be formulated by incorporating a discontinuous term
into $\hat{u}$ in Equation (\ref{u_hat_equ}) as follows: 
\begin{equation}
	u=\hat{u}+\frac{1}{b}\begin{bmatrix}\lambda_{1}S+\lambda_{2}sgn(S)\end{bmatrix}\label{control_final1_equ}
\end{equation}
where $\lambda_{1}>0$ and $\lambda_{2}>0$.

\subsection{Stability Analysis}
\begin{thm}
	\label{them1}Consider the system dynamics in (\ref{xsi_equ}) and
	the proposed NFTSMC surfaces described in Equation (\ref{sliding_surface_equ}).
	The proposed controller designed presented in Equation (\ref{control_final1_equ})
	ensures asymptotic stability of the closed loop system in the sense
	of Lyapunov. 
\end{thm}
\textbf{Proof}: to prove the stability of the system using the Lyapunov
method. Let $V$ be a Lyapunov function candidate for the system as
follows: 
\begin{equation}
	V=\frac{1}{2}S^{2}\label{lyap_fun_equ}
\end{equation}
where $V=0$ when $S=0$, and $V>0,~\forall~S\neq0$, and its time
derivative exists and given as follows: 
\begin{equation}
	\dot{V}=S\dot{S}\label{vdot_equ}
\end{equation}
Substituting Equation (\ref{sdot_equ}) into Equation (\ref{vdot_equ})
results in the following: 
\begin{equation}
	\dot{V}=\begin{bmatrix}\ddot{\chi_{d}}-\digamma-\delta-bu+(\xi+\eta a|e|^{a-1})\dot{e}\end{bmatrix}S\label{vsdot_equ}
\end{equation}
Substituting Equation (\ref{control_final1_equ}) into Equation (\ref{vsdot_equ})
results in the following: 
\begin{equation}
	\begin{split}\dot{V} & =\begin{bmatrix}-\lambda_{1}S-\lambda_{2}sgn(S)\end{bmatrix}S\\
		& =-\lambda_{1}S^{2}-\lambda_{2}|S|
	\end{split}
	\label{vsdot_f_equ}
\end{equation}
Since $\lambda_{1}>0\implies\lambda_{1}S^{2}>0~,\forall~S\neq0$,~
Also, $\lambda_{2}>0\implies\lambda_{2}|S|>0~,\forall~S\neq0$. Therefore,
$\dot{V}<0~,\forall~S\neq0\implies\dot{V}\leqslant0~,\forall~S$ which
satisfies the Lyapunov stability condition and implies that the states
will converge to the NFTSM surfaces $(S=0)$ as well as the errors
will asymptotically converge to the origin. 
\begin{lem}
	\label{lem_time}Using proposed controller in Equation (\ref{control_final1_equ}),
	the system trajectories can reach the equilibrium state on the NFTSM
	sliding surface $(S=0)$ within a finite time. 
\end{lem}
\textbf{Proof}: to prove that the states of the system converge to
the NFTSM surfaces $(S=0)$ in a finite time, lets use the result
of Theorem \ref{them1} in Equation (\ref{vsdot_f_equ}) to derive
the reaching time $(t_{r})$ as follows: 
\begin{equation}
	\dot{V}=\frac{dV}{dt}=-\lambda_{1}S^{2}-\lambda_{2}|S|\label{vdot_tr_equ}
\end{equation}
Using Equation (\ref{lyap_fun_equ}), Equation (\ref{vdot_tr_equ})
can be rewritten as follows: 
\[
\frac{dV}{dt}\leqslant-2\lambda_{1}V-\lambda_{2}\sqrt{2V},~~~\text{such that:}
\]
\begin{equation}
	dt\leqslant\frac{-dV}{2\lambda_{1}V+\gamma V^{\frac{1}{2}}}\label{equ_dt}
\end{equation}
Where $\gamma=\sqrt{2}\lambda_{2}$, then integrate both sides of
Equation (\ref{equ_dt}) such that: 
\begin{equation}
	\int_{0}^{t_{r}}dt\leqslant\int_{V(0)}^{V(t_{r})}\frac{-dV}{2\lambda_{1}V+\gamma V^{\frac{1}{2}}}\label{equ_dt_int}
\end{equation}
Using substitution to integrate the right side of Equation (\ref{equ_dt_int}),
we obtain: 
\begin{equation}
	\begin{split}\int_{V(0)}^{V(t_{r})}\frac{-dV}{2\lambda_{1}V+\gamma V^{\frac{1}{2}}} & =-\frac{1}{\lambda_{1}}\Big[\ln|2\lambda_{1}V^{\frac{1}{2}}+\gamma|\Big]_{V(0)}^{V(t_{r})}\\
		& =\frac{1}{\lambda_{1}}\ln\left|\frac{2\lambda_{1}V(0)^{\frac{1}{2}}+\gamma}{2\lambda_{1}V(t_{r})^{\frac{1}{2}}+\gamma}\right|
	\end{split}
	\label{equ_integral}
\end{equation}
Integrating the left side of Equation (\ref{equ_dt_int}) and using
the result in Equation (\ref{equ_integral}), the reaching time $(t_{r})$
is given as: 
\begin{equation}
	t_{r}\leqslant\frac{1}{\lambda_{1}}\ln\left|\frac{2\lambda_{1}V(0)^{\frac{1}{2}}+\gamma}{\gamma}\right|\label{reach_time_equ}
\end{equation}
Based on (\ref{reach_time_equ}), Lemma \ref{lem_time} is proven
and he system trajectories will reach (S = 0) within a finite time.

\subsection{Position Controller}

Using the controller designed in Equation (\ref{control_final1_equ}),
the virtual control inputs can be expressed as follows: 
\begin{equation}
	\begin{cases}
		u_{x}= & m_{t}\ddot{x_{d}}+k_{l_{{drag}_{x}}}\dot{x}-d_{l_{1}}\\
		& +m_{t}\left((\xi+\eta a|e_{x}|^{a-1})\dot{e}_{x}+\lambda_{1}S_{x}+\lambda_{2}sgn(S_{x})\right)\\
		u_{y}= & \ddot{y_{d}}+k_{l_{{drag}_{y}}}\dot{y}-d_{l_{2}}\\
		& +m_{t}\left((\xi+\eta a|e_{y}|^{a-1})\dot{e}_{y}+\lambda_{1}S_{y}+\lambda_{2}sgn(S_{y})\right)\\
		u_{z}= & m_{t}\ddot{z_{d}}+m_{t}g+k_{l_{{drag}_{z}}}\dot{z}-d_{l_{3}}\\
		& +(\xi+\eta a|e_{z}|^{a-1})\dot{e}_{z}+\lambda_{1}S_{z}+\lambda_{2}sgn(S_{z})]
	\end{cases}\label{pos_cont_equ}
\end{equation}
According to Equations (\ref{vertual_cont_equ}) and (\ref{pos_cont_equ}),
the total thrust $U_{1}$ is given as follows: 
\begin{align}
	U_{1}= & \frac{m_{t}}{c\phi c\theta}\left(\ddot{z_{d}}+g+\frac{1}{m_{t}}k_{l_{{drag}_{z}}}\dot{z}-\frac{d_{l_{3}}}{m_{t}}\right)\nonumber \\
	& +\frac{m_{t}}{c\phi c\theta}\left((\xi+\eta a|e_{z}|^{a-1})\dot{e}_{z}+\lambda_{1}S_{z}+\lambda_{2}sgn(S_{z})\right)\label{thrus_z_equ}
\end{align}
and the desired roll and pitch angles $(\phi_{d}$ and $\theta_{d})$
can be calculated by utilizing the tangent function within limits
$(-\frac{\pi}{2},~\frac{\pi}{2})$ of Euler angles, as follows: 
\begin{equation}
	\begin{aligned}\theta_{d} & =\arctan\Biggl(\frac{u_{x}\cos(\psi_{d})+u_{y}\sin(\psi_{d})}{u_{z}}\Biggl),\\
		\phi_{d} & =\arctan\Biggl(\cos(\theta_{d})\Bigl(\frac{u_{x}\sin(\psi_{d})-u_{y}\cos(\psi_{d})}{u_{z}}\Bigl)\Biggl)
	\end{aligned}
	\label{equ_des_orient}
\end{equation}

\subsection{Attitude Controller}

The attitude controller takes the desired roll and pitch angles $(\phi_{d}$
and $\theta_{d})$ that are generated by the position controller in
Equation (\ref{equ_des_orient}) with the desired yaw angle $\psi_{d}$
which is set by the designer according to the configuration of entire
system to calculate the rolling, pitching, and yawing moments according
to the general controller law in Equation (\ref{control_final1_equ})
as follows: 
\begin{align}
	U_{2}= & J_{x}\left(\ddot{\phi}_{d}-\frac{(J_{y}-J_{z})}{J_{x}}\dot{\theta}\dot{\psi}+\frac{k_{r_{{drag}_{\phi}}}}{J_{x}}\dot{\phi}-\frac{d_{r_{1}}}{J_{x}}\right)\nonumber \\
	& +J_{x}\left((\xi+\eta a|e_{\phi}|^{a-1})\dot{e}_{\phi}+\lambda_{1}S_{\phi}+\lambda_{2}sgn(S_{\phi})\right)\nonumber \\
	U_{3}= & J_{y}\left(\ddot{\theta}_{d}-\frac{(J_{z}-J_{x})}{J_{y}}\dot{\phi}\dot{\psi}+\frac{k_{r_{{drag}_{\theta}}}}{J_{y}}\dot{\theta}-\frac{d_{r_{2}}}{J_{y}}\right)\nonumber \\
	& +J_{y}\left((\xi+\eta a|e_{\theta}|^{a-1})\dot{e}_{\theta}+\lambda_{1}S_{\theta}+\lambda_{2}sgn(S_{\theta})\right)\nonumber \\
	U_{4}= & J_{z}\left(\ddot{\psi}_{d}-\frac{(J_{x}-J_{y})}{J_{z}}\dot{\phi}\dot{\theta}+\frac{k_{r_{{drag}_{\psi}}}}{J_{z}}\dot{\psi}-\frac{d_{r_{3}}}{J_{z}}\right)\nonumber \\
	& +J_{z}\left((\xi+\eta a|e_{\psi}|^{a-1})\dot{e}_{\psi}+\lambda_{1}S_{\psi}+\lambda_{2}sgn(S_{\psi})\right)\label{orient_cont_equ}
\end{align}

\subsection{Chattering Problem}

Chattering occurs due to the discontinuous nature of the control law
in sliding mode control (SMC). The control signal rapidly switches
between two values, causing high-frequency oscillations around the
sliding surface. Chattering can lead to mechanical wear, instability,
and poor performance of actuators. Using sign function causes the
chattering problem in the control signal which is undesirable for
the aforementioned reasons and needs to be eliminated. To achieve
this, the control discontinuity can be smoothed out in a thin boundary
layer in vicinity of the switching surface by using hyperbolic tangent
function $tanh$ or saturation function $sat$ which is given in Equation
(\ref{sat_equ}) as follows: 
\begin{equation}
	sat(S):=\begin{cases}
		1 & \text{if }S>\Phi,\\
		\frac{S}{\Phi} & \text{if }|S|\leqslant\Phi,\\
		-1 & \text{if }S<\Phi
	\end{cases}\label{sat_equ}
\end{equation}
where $0<\Phi<1$ represents the thin boundary layer. Modifying the
controllers in Equations (\ref{thrus_z_equ}) and (\ref{orient_cont_equ}),
the chattering-free controllers are given as follows: 
\begin{align}
	U_{1}= & \frac{m_{t}}{c\phi c\theta}\left(\ddot{z_{d}}+g+\frac{1}{m_{t}}k_{l_{{drag}_{z}}}\dot{z}-\frac{d_{l_{3}}}{m_{t}}\right)\nonumber \\
	& +\frac{m_{t}}{c\phi c\theta}\left((\xi+\eta a|e_{z}|^{a-1})\dot{e}_{z}+\lambda_{1}S_{z}+\lambda_{2}sat(S_{z}/\Phi)\right)\nonumber \\
	U_{2}= & J_{x}\left(\ddot{\phi}_{d}-\frac{(J_{y}-J_{z})}{J_{x}}\dot{\theta}\dot{\psi}+\frac{k_{r_{{drag}_{\phi}}}}{J_{x}}\dot{\phi}-\frac{d_{r_{1}}}{J_{x}}\right)\nonumber \\
	& +J_{x}\left((\xi+\eta a|e_{\phi}|^{a-1})\dot{e}_{\phi}+\lambda_{1}S_{\phi}+\lambda_{2}sat(S_{\phi}/\Phi)\right)\nonumber \\
	U_{3}= & J_{y}\left(\ddot{\theta}_{d}-\frac{(J_{z}-J_{x})}{J_{y}}\dot{\phi}\dot{\psi}+\frac{k_{r_{{drag}_{\theta}}}}{J_{y}}\dot{\theta}-\frac{d_{r_{2}}}{J_{y}}\right)\nonumber \\
	& +J_{y}\left((\xi+\eta a|e_{\theta}|^{a-1})\dot{e}_{\theta}+\lambda_{1}S_{\theta}+\lambda_{2}sat(S_{\theta}/\Phi)\right)\nonumber \\
	U_{4}= & J_{z}\left(\ddot{\psi}_{d}-\frac{(J_{x}-J_{y})}{J_{z}}\dot{\phi}\dot{\theta}+\frac{k_{r_{{drag}_{\psi}}}}{J_{z}}\dot{\psi}-\frac{d_{r_{3}}}{J_{z}}\right)\nonumber \\
	& +J_{z}\left((\xi+\eta a|e_{\psi}|^{a-1})\dot{e}_{\psi}+\lambda_{1}S_{\psi}+\lambda_{2}sat(S_{\psi})/\Phi\right)\label{control_final_equ}
\end{align}

\subsection{Admittance Controller \label{controller_design}}

\begin{figure}
	\centering{}\includegraphics[scale=0.33]{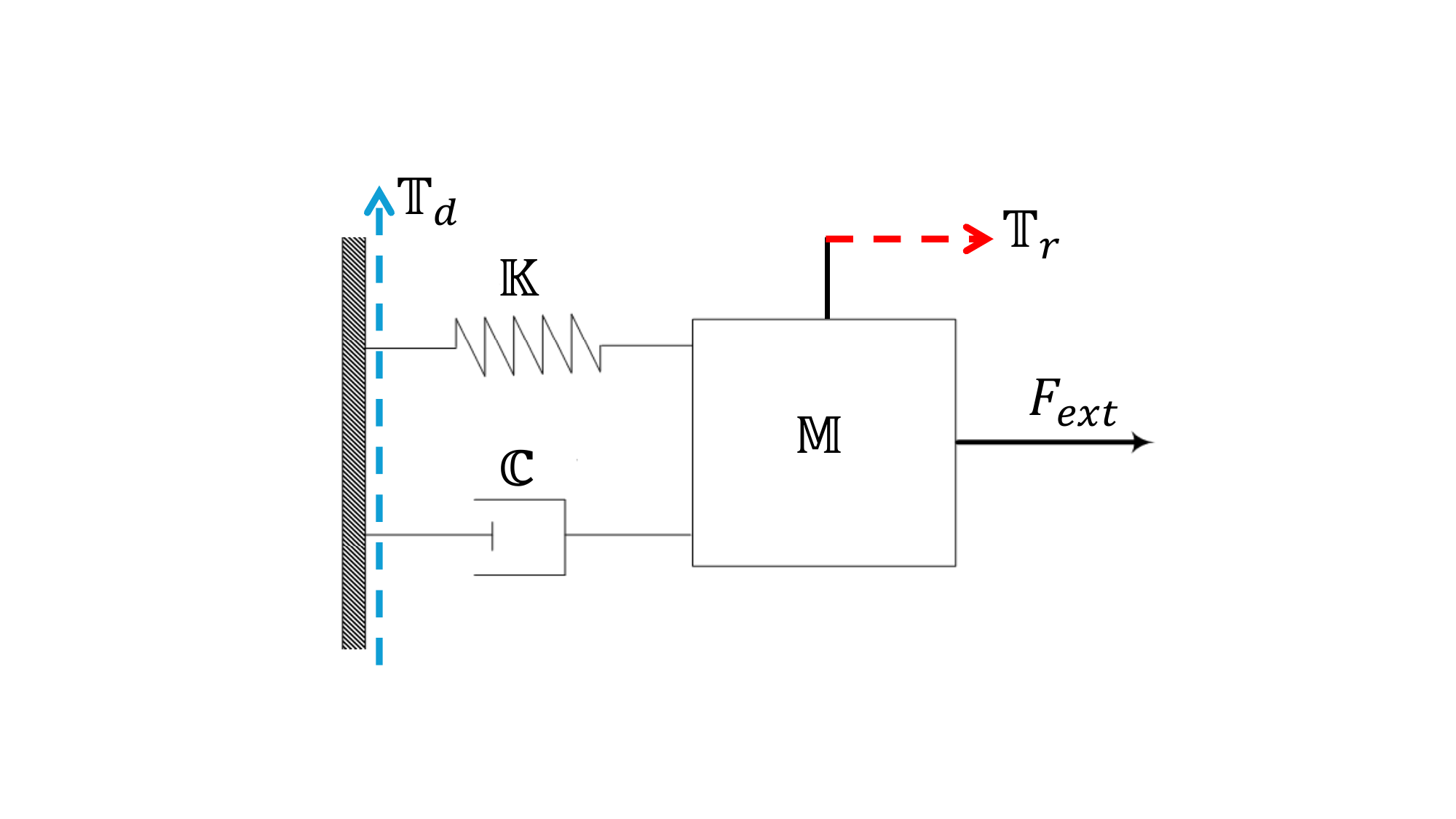}\caption{\label{fig3}Virtual mass-damper-spring system.}
\end{figure}

Admittance controller is widely utilized in robotics and control systems
\cite{ref17,ref30,ref32}, enabling a robot or a system to respond
effectively to external forces or disturbances. It provides the robot
with a degree of ``softness'' or ``compliance'', facilitating
interaction with human and its environment in a safe and adaptive
manner. In essence, the term ``admittance'' originates from electrical
engineering, describing how a circuit reacts to an input voltage.
In the realm of robotics, it has been adapted to characterize how
a system reacts to external forces. The admittance controller enables
attaining the desired interaction dynamics by adjusting the robot's
movement based on the measured or estimated interaction force. This
control method governs the dynamic interaction between the motion
parameters (such as position, velocity, and acceleration) and the
input force by modifying the virtual inertia, damping, and stiffness
of the robot. It simulates the behavior of the mass-damper-spring
system through the mathematical relationship defined in Equation (\ref{equ4}).
Figure \ref{fig3} depicts the behavior of the system as a mass-damper-spring
system. 
\begin{equation}
	\mathbb{M}(\ddot{\mathbb{T}}_{d}-\ddot{\mathbb{T}}_{r})+\mathbb{C}(\dot{\mathbb{T}}_{d}-\dot{\mathbb{T}}_{r})+\mathbb{K}(\mathbb{T}_{d}-\mathbb{T}_{r})=F_{ext}\label{equ4}
\end{equation}
where: $\mathbb{T}_{d}\in\mathbb{R}^{3}$ is the desired trajectory,
$\mathbb{T}_{r}\in\mathbb{R}^{3}$ is the reference trajectory generated
by the admittance controller according to human guidance proportional
to the applied forces, $\mathbb{M}\in\mathbb{R}^{3\times3},~\mathbb{C}\in\mathbb{R}^{3\times3},~\mathbb{K}\in\mathbb{R}^{3\times3}$
are the diagonal matrices of the virtual mass, damping, stiffness
of the system, and $F_{ext}\in\mathbb{R}^{3}$ is the vector of external
forces acting on the system due to human physical interaction and
guidance. All quantities are expressed with respect to the inertial
reference frame. 

\begin{figure*}
	\centering{}\includegraphics[width=1\textwidth]{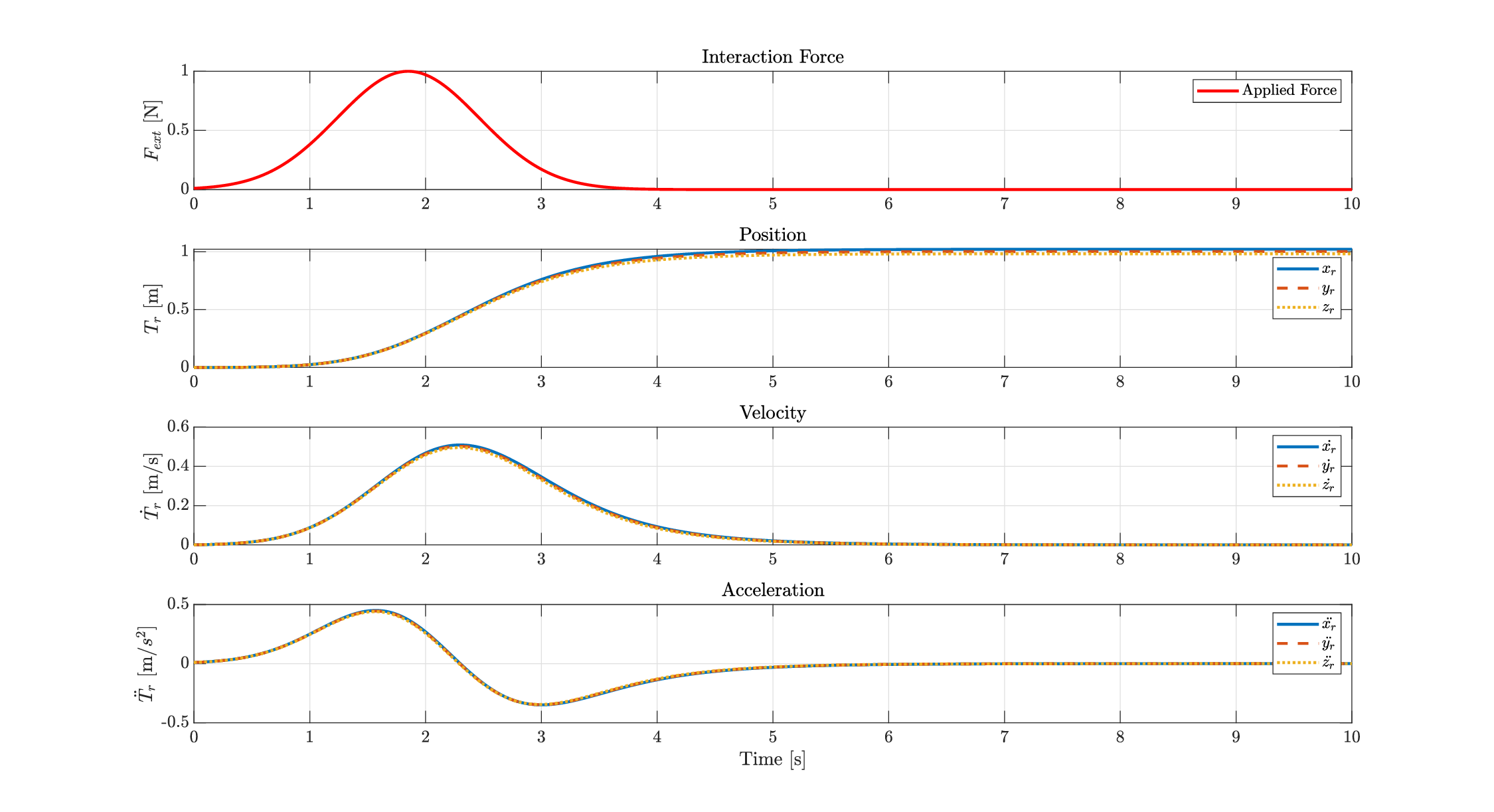}\caption{\label{fig4}Reference trajectories generated by the admittance controller
		proportional to the applied force.}
\end{figure*}

In this work, the admittance controller is designed to enable seamless
interaction between human operator and aerial systems. It adjusts
the aerial vehicle's behavior in real-time based on the force exerted
by the human operator, allowing for precise control and coordination
during payload transportation tasks. It generates a new reference
trajectory proportional to the amount of the applied force as illustrated
in Figure \ref{fig4}. The reference trajectories in Figure \ref{fig4}
represent the output of the admittance controller in response to the
applied force in the $(x,y,z)$ directions. The applied force is Gaussian-shaped,
which is normalized in this graph. 

The response of the physical interaction can be adapted by tuning
the virtual spring stiffness constant $\mathbb{K}$. Increasing the
value of $\mathbb{K}$ produces a stiffer response while setting the
value of $\mathbb{K}$ to zero guarantees full compliance with the
applied external force. The desired acceleration $\ddot{\mathbb{T}}_{d}$
and velocity $\dot{\mathbb{T}}_{d}$ are set to zero to enforce the
system to maintain its position as long as no interaction force is
applied. The controller architecture is optimized to ensure stability,
robustness, and responsiveness in dynamic environments. In order to
reject any environmental disturbances, the admittance controller response
is triggered if the measured force exceeds a specific threshold.

\section{Numerical Results and Simulation\label{results-discussion}}

\subsection{Simulation Setup and Implementation}

The simulation environment is configured to replicate real-world conditions
and scenarios encountered during payload transportation tasks and
physical interaction. In this section, we set up the simulation environment
of the proposed assistive payload transportation system with human
guidance thorough physical interaction. The system consists of two
quadrotors and a common payload. The quadrotors are identical with
known geometrical and physical specifications. The payload is a uniform
beam-shaped circular bar. All components of the system are rigidly
connected, where each quadrotor is connected to each end of the payload.
There are two force-torque sensors at the points of contact of rigid
connection between the quadrotors and the payload to measure the interaction
and reaction forces and torques between system's components as well
as the human-robot physical interaction. The system operates near
hovering state with small attitude angles. When the human operator
applies force on the payload, the force will be measure by the force-torque
sensors and send to the admittance controller. The latter regulates
the virtual parameters of the system and generates a reference trajectory
for the system to follow proportional to the amount and direction
of the measured force according to human guidance. The reference trajectory
is delivered to the position controller as a desired trajectory to
generate the required thrust, as well as the desired orientation to
the attitude controller to achieve the required movement and follow
the human guidance with full compliance.

\begin{figure}
	
	\begin{centering}
		\subfloat[]{
			\begin{centering}
				\includegraphics[scale=0.7]{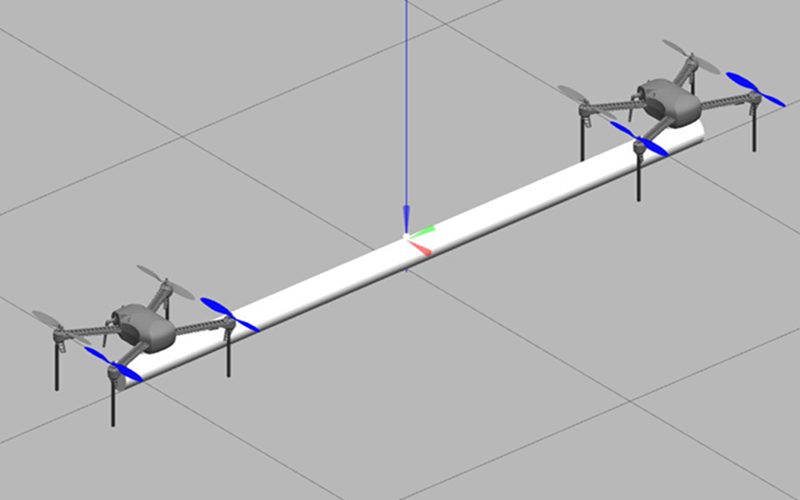}
				\par\end{centering}
			
		}
		\par\end{centering}
	\centering{}\subfloat[]{\begin{centering}
			\includegraphics[scale=0.6]{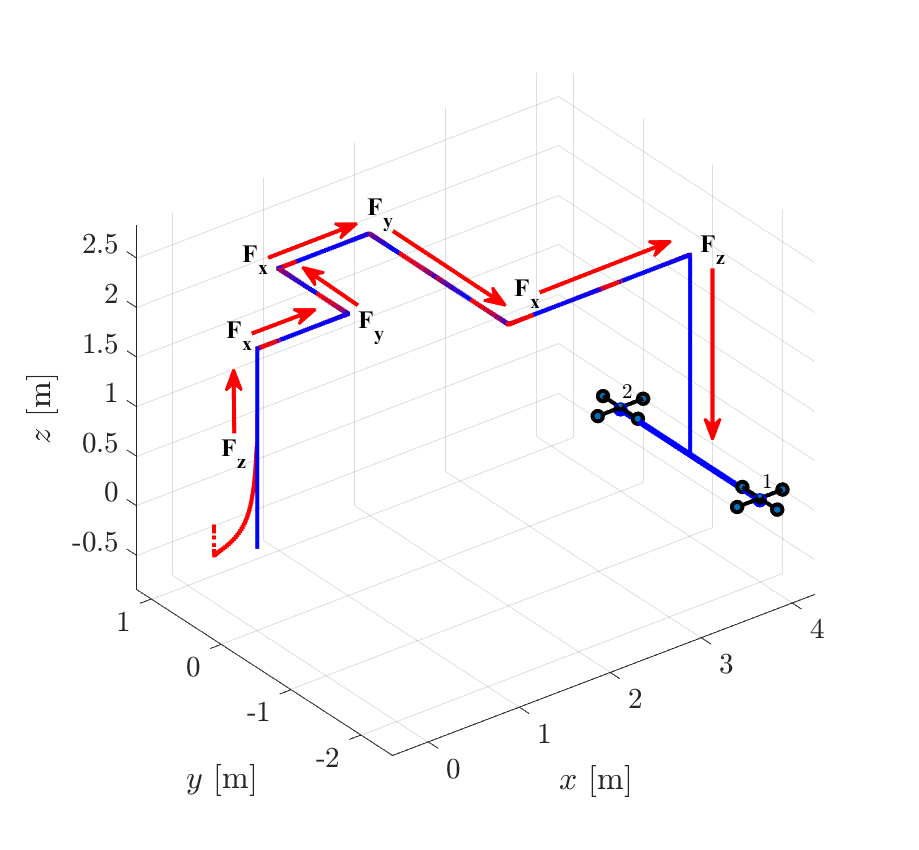}
			\par\end{centering}
		
	}\caption{\label{fig5}The simulation environment setup; (a) The entire aerial
		vehicle built in ROS and Gazebo environment, (b) MATLAB simulation
		of the assistive system's motion following human guidance in 3D space
		as depicted by the red arrows.}
\end{figure}

MATLAB is used as a simulation environment. The simulation is implemented
to effectively assess the efficiency and performance of the control
strategy suggested in Section \ref{control_section} and stated in
Equation (\ref{control_final_equ}). All simulation parameters, including
individual quadrotor parameters, payload, the entire assistive payload
transportation system, and the parameters of the position, attitude,
and admittance controllers are given in Tables \ref{tab:entire_sys_parameters}
and \ref{tab:controller_parameters}, respectively. Additionally,
the simulation environment is implemented using the Robotic Operation
System (ROS noetic) and Gazebo 11, providing a realistic and flexible
platform for experimentation. An IRIS quadrotor model was used to
simulate the quadrotors and built the entire aerial vehicle. Figure
\ref{fig5}.(a) shows the entire aerial vehicle built in Gazebo11
environment while Figure \ref{fig5}.(b) shows the MATLAB simulation
of guided motion in $3D$ space of the assistive payload transportation
system according to the human guidance and physical interaction. The
human operator can apply forces on the payload in any direction to
move the system to the direction of the applied force. 
\begin{table}[t]
	\centering{}\caption{\label{tab:entire_sys_parameters} System parameters.}
	\begin{tabular}{llr}
		\hline 
		Symbol  & Definition  & Value/ Unit\tabularnewline
		\hline 
		\hline 
		$m_{i}$  & Mass of $i^{th}$ quadrotor  & $1.5/kg$\tabularnewline
		$J_{x}$  & Moment of inertia of the x-axis  & $2.9125\times10^{-2}/kg~m^{2}$\tabularnewline
		$J_{y}$  & Moment in inertia of the y-axis  & $2.9125\times10^{-2}/kg~m^{2}$\tabularnewline
		$J_{z}$  & Moment in inertia of the z-axis  & $5.5225\times10^{-2}/kg~m^{2}$\tabularnewline
		$l$  & Arm length of $i^{th}$ quadrotor  & $0.25/m$\tabularnewline
		$m_{L}$  & Mass  & $0.5/kg$\tabularnewline
		$J_{Lx}$  & Moment of inertia of the x-axis  & $16.6667\times10^{-2}/kg~m^{2}$\tabularnewline
		$J_{Ly}$  & Moment in inertia of the y-axis  & $6.25\times10^{-4}/kg~m^{2}$\tabularnewline
		$J_{Lz}$  & Moment in inertia of the z-axis  & $16.6667\times10^{-2}/kg~m^{2}$\tabularnewline
		$L$  & Length  & $2/m$\tabularnewline
		$r_{L}$  & Radius  & $0.05/m$\tabularnewline
		$m_{t}$  & Total mass of entire system  & $3.5/kg$\tabularnewline
		$g$  & Gravitational acceleration  & $9.81/m/s^{2}$\tabularnewline
		$J_{tx}$  & Moment of inertia of the x-axis  & $3.227327/kg~m^{2}$\tabularnewline
		$J_{ty}$  & Moment of inertia of the y-axis  & $0.061286/kg~m^{2}$\tabularnewline
		$J_{tz}$  & Moment of inertia of the z-axis  & $3.277117/kg~m^{2}$\tabularnewline
		$k_{l_{{drag}_{x}}}$  & Drag coefficients along the x-axis  & $6\times10^{-3}/N~s/m$\tabularnewline
		$k_{l_{{drag}_{y}}}$  & Drag coefficients along the y-axis  & $6\times10^{-3}/N~s/m$\tabularnewline
		$k_{l_{{drag}_{z}}}$  & Drag coefficients along the z-axis  & $6\times10^{-3}/N~s/m$\tabularnewline
		$k_{r_{{drag}_{\phi}}}$  & Drag coefficients about the x-axis  & $6\times10^{-3}/N~s$\tabularnewline
		$k_{r_{{drag}_{\theta}}}$  & Drag coefficients about the y-axis  & $6\times10^{-3}/N~s$\tabularnewline
		$k_{r_{{drag}_{\psi}}}$  & Drag coefficients about the z-axis  & $6\times10^{-3}/N~s$\tabularnewline
		\hline 
	\end{tabular} 
\end{table}

\begin{table}[t]
	\centering{}\centering \caption{\label{tab:controller_parameters} Parameters of the Controller.}
	\begin{tabular}{llr}
		\hline 
		Symbol  & Definition  & Value/ Unit\tabularnewline
		\hline 
		\hline 
		$\xi_{x},\xi_{y},\xi_{z}$  & Control constants  & 4, 2, 11\tabularnewline
		$\xi_{\phi},\xi_{\theta},\xi_{\psi}$  & Control constants  & 25, 80, 25\tabularnewline
		$\eta_{x},\eta_{y},\eta_{z}$  & Control constants  & 0.2, 0.1, 0.2\tabularnewline
		$\eta_{\phi},\eta_{\theta},\eta_{\psi}$  & Control constants  & 0.2, 0.2, 0.2\tabularnewline
		$\lambda_{1x},\lambda_{1y},\lambda_{1z}$  & Control constants  & 0.2, 0.1, 200\tabularnewline
		$\lambda_{2x},\lambda_{2y},\lambda_{2z}$  & Control constants  & 2, 1, 100\tabularnewline
		$\lambda_{1_{\phi}},\lambda_{1_{\theta}},\lambda_{1_{\psi}}$  & Control constants  & 40, 40, 40\tabularnewline
		$\lambda_{2_{\phi}},\lambda_{2_{\theta}},\lambda_{2_{\psi}}$  & Control constants  & 30, 30, 30\tabularnewline
		$a$  & Positive constant  & 3\tabularnewline
		$\mathbb{M}$  & Virtual mass  & $1/kg$\tabularnewline
		$\mathbb{C}$  & Virtual damping coefficients  & $1.6/N~s/m$\tabularnewline
		$\mathbb{K}$  & Virtual spring constants  & $0/N~s/m$\tabularnewline
		\hline 
	\end{tabular}
\end{table}

\subsection{Numerical Results and Discussion}

\begin{figure*}
	
	\begin{centering}
		\includegraphics[width=1\textwidth]{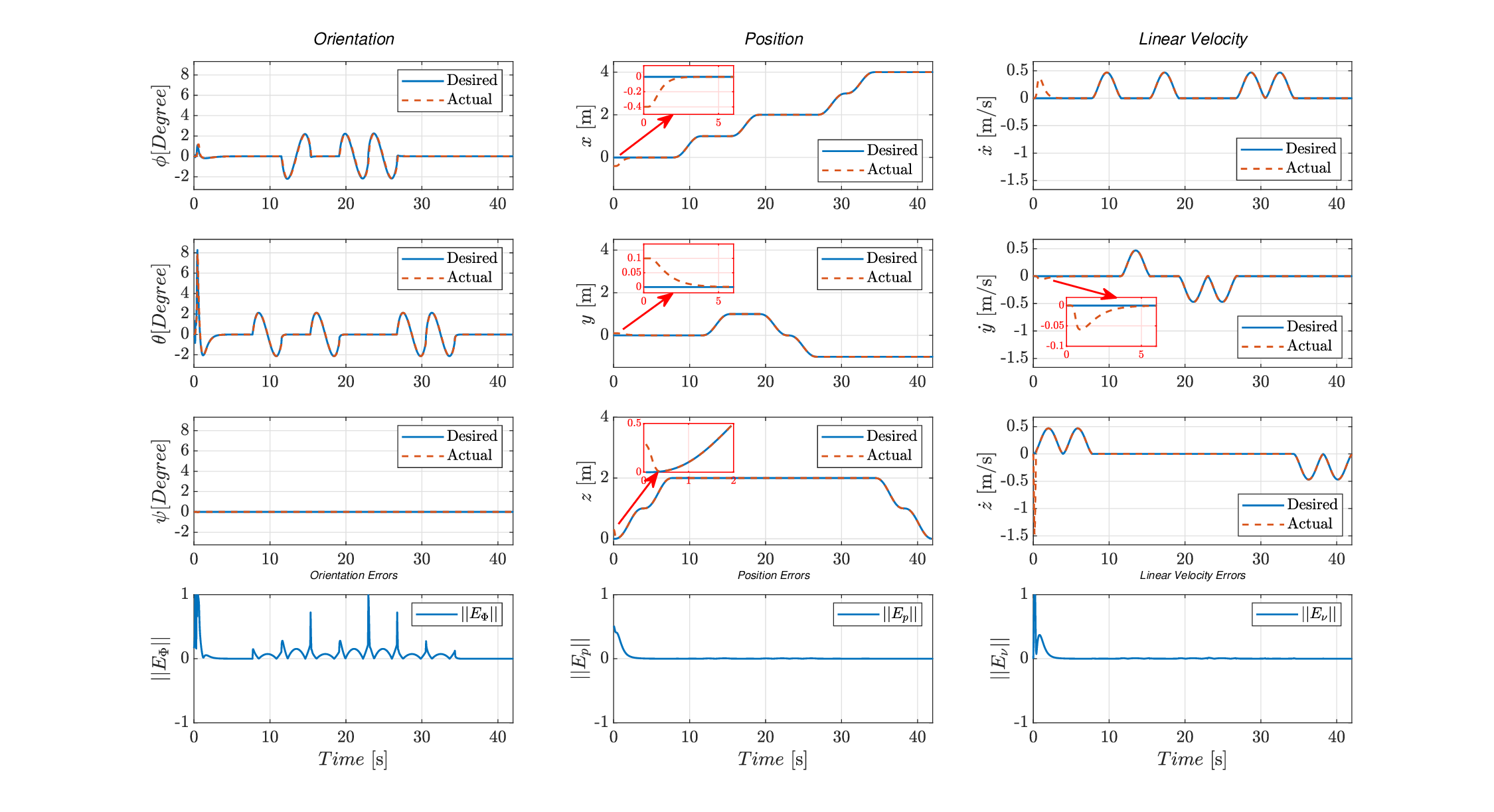}
		\par\end{centering}
	\caption{\label{fig6}The simulation results; First column for the desired
		and actual orientation of the system, Second column for the desired
		and actual position, Third column for the linear velocity of the system
		in $(x,y,z)$ directions, and the bottom row for the normalized tracking
		errors of the system (orientation, position, and linear velocity).}
\end{figure*}

Extensive simulations were conducted to evaluate the performance of
the proposed assistive payload transportation system with human physical
interaction using quadrotors, as well as to assess the effectiveness
of the proposed controller to stabilize the system and track the desired
human guidance. The results indicate that the integration of human
operators into the control loop of the system using the admittance
controller significantly improves the payload transport capabilities.
The system exhibits enhanced adaptability, flexibility, and safety,
making it suitable for a wide range of applications in dynamic and
unpredictable environments. At first, the aerial vehicle lifted the
payload to a predefined height. At this moment, the admittance controller
was engaged to be ready for physical interaction. As it is illustrated
by red arrows in Figure \ref{fig5}.(b), a force was applied upward
to place the aerial vehicle at an appropriate operating height. Then
a simulated human guidance was performed by applying forces in different
directions to transport the payload to the final destination. The
results showed full compliance of the system with the applied human
guidance, while the controller maintained the stability of the system
during the process until the final destination. The results of the
NFTSMC position and attitude controllers are illustrated as follows:

Figure \ref{fig6} shows the performance results of the system and
its trajectories including the desired and actual orientation, position,
velocity, and tracking errors. The actual trajectories tracked the
reference trajectories that were generated by the admittance controller
as desired trajectories for the position and attitude controllers.
One can see that the controllers perfectly tracked the desired trajectories
and followed the human operator's guidance. Additionally, the results
show that the controllers were able to stabilize the system and reduced
the errors to zero after an aggressive start at the beginning of the
simulation. The first column of the figure shows the desired and actual
orientation of the system $(\phi,\theta,\psi)$. It can be clearly
seen that the position controller received the reference trajectories
from the admittance controller and managed to generate the required
thrust and desired orientation $(\phi_{d})$ and $(\theta_{d})$.
The desired orientation were delivered to the attitude controller
while the latter tracked them to move the system to the positions
in the $x$ and $(y)$ directions according to human guidance. Notice
that the $(\psi)$ angle is kept at zero as it desired by the designer.
The second column shows the corresponding trajectories of the system
in the $(x,y,z)$ directions. The $x$ graph shows the forward direction
of the system which gradually increased following the applied forces
alongside with the velocity profile which went up and down for each
movement, depicting the required increase in the velocity at the beginning
of the movement and the decrease in the velocity as the system prepares
for full stop waiting for upcoming physical interaction. In addition,
the $(y)$ direction graph shows that the system went left and right,
navigating according to human guidance and the applied physical interaction
force.

Moreover, the $(z)$ direction graph shows that the altitude of the
system increased to the appropriate operating height and maintained
the same altitude until it reached the destination and finally landed
while the velocity graph shows the related velocity profile. Furthermore,
Figure \ref{fig6} shows the normalized tracking errors of the system
for the the orientation, position, and linear velocity, respectively.
It can be noticed the errors of position and the velocity were driven
to zero and kept at zero during the entire mission time, indicating
the robust performance of the proposed controller. On the other hand,
the orientation error shows some fluctuations which are related to
the high rolling and pitching control inputs at the start and stop
points of the system due to high inertia, however the error is bounded
and driven to zero by the controller at each point.

\begin{figure}[htbp]
	\centering{}\includegraphics[width=0.85\linewidth]{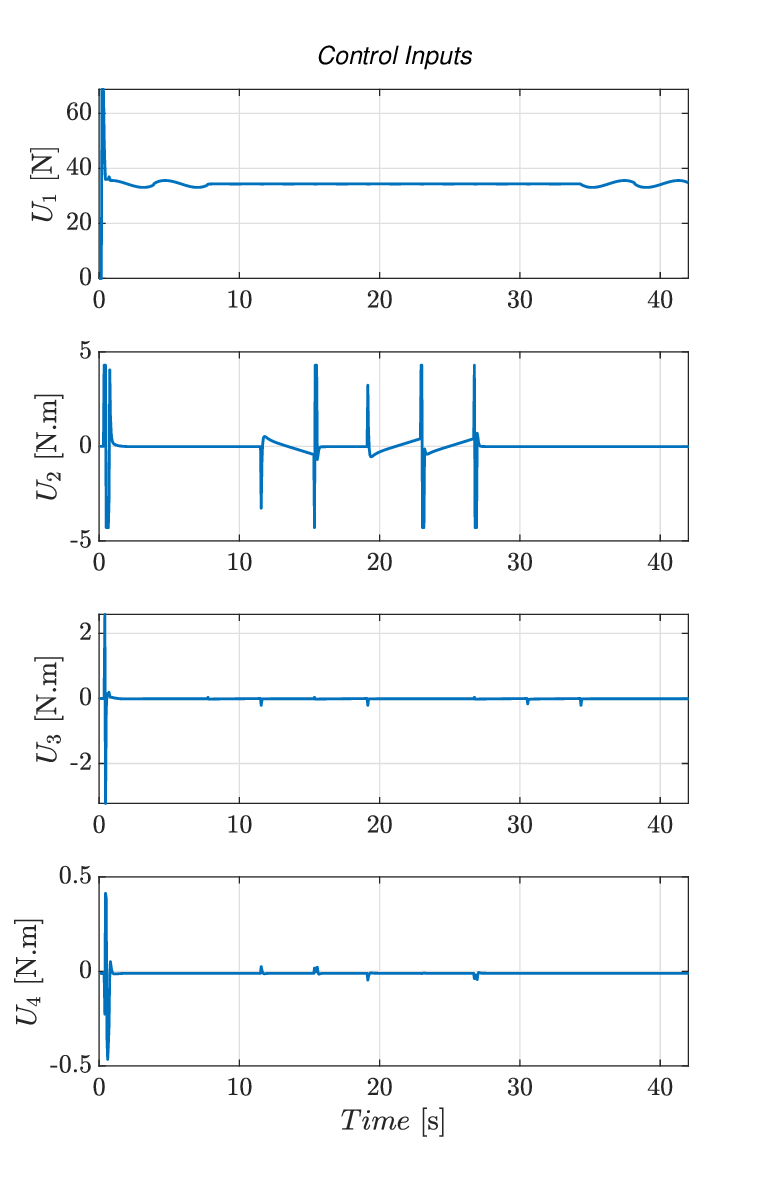}
	\caption{\label{fig7}The control inputs.}
\end{figure}

\begin{figure}[htbp]
	\centering{}\includegraphics[width=0.8\linewidth]{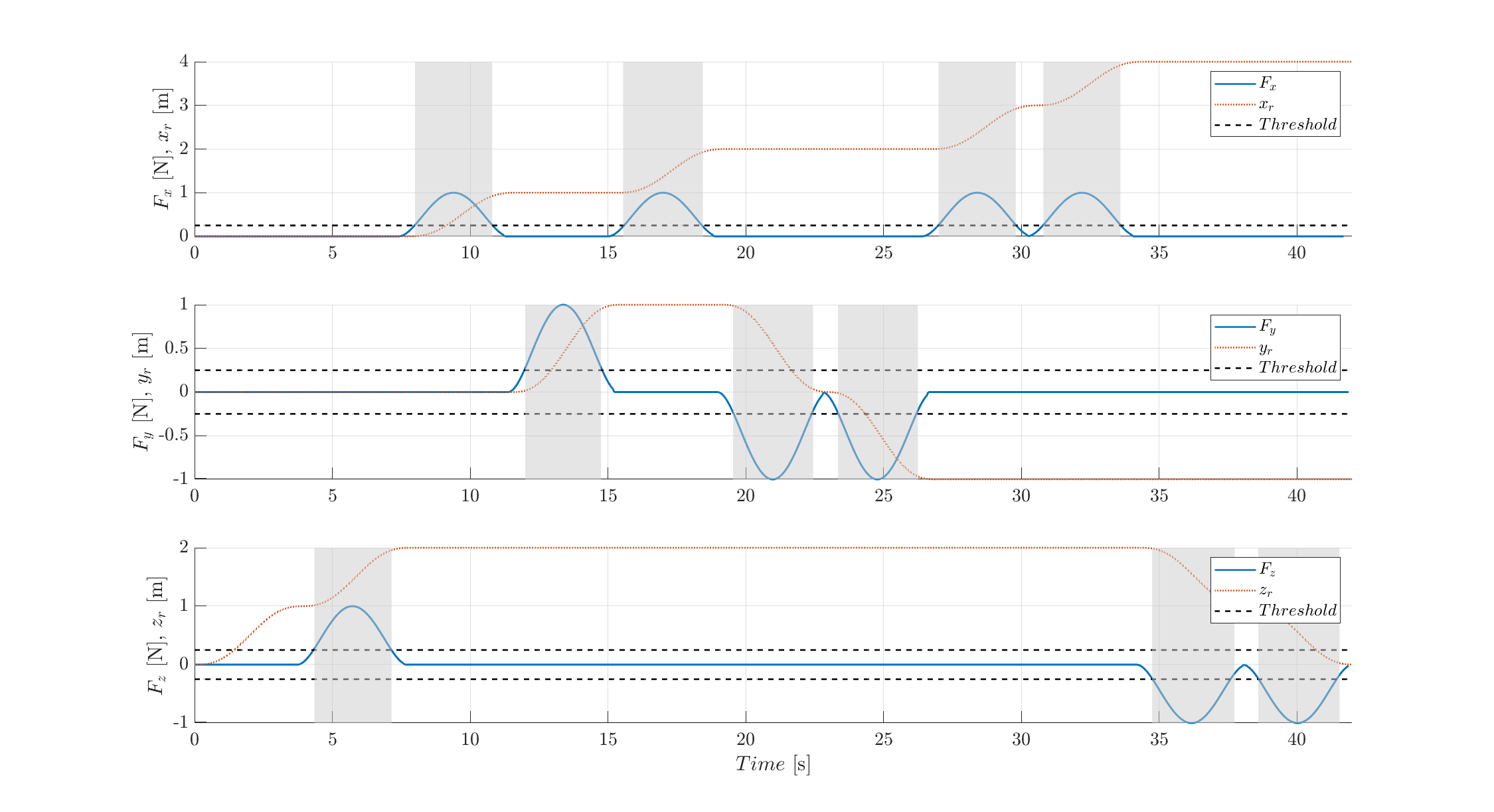}
	\caption{\label{fig8}The human interaction with the system in a 3D environment
		is depicted. The plots illustrate the external forces applied (solid
		blue) to the center of mass (CoM) of the payload, along with the system's
		reference trajectories (dotted red) in the x, y, and z directions,
		respectively. The dashed black lines represent the thresholds that
		regulate the transition between the admittance controller (gray shaded
		areas) and the position controller.}
\end{figure}

Figure \ref{fig7} shows the calculated control inputs that include
the total thrust, rolling, pitching, and yawing moments calculated
according to the proposed NFTSMC position and attitude controllers.
It can be seen that the controllers took appropriate actions to stabilize
the system at the disturbed starting point in the beginning of the
simulation. Then, they returned back to their normal rhythm to generate
the control signals making the system performed appropriate navigation
under the simulated human guidance. The human guidance and interaction
with the system in a 3D environment is depicted in Figure \ref{fig8}.
The plots illustrate the external forces applied (solid blue) on the
center of mass (CoM) of the payload, alongside with the system's reference
trajectories (dotted red) in the x, y, and z directions. The dashed
black lines represent the thresholds that regulate the transition
between the actions of the admittance controller (gray-shaded areas)
and the position controller (white areas) which work together to achieve
the transportation task. 

\begin{figure*}[hbt!]
	\begin{centering}
		\subfloat[]{\begin{centering}
				\includegraphics[width=0.45\linewidth]{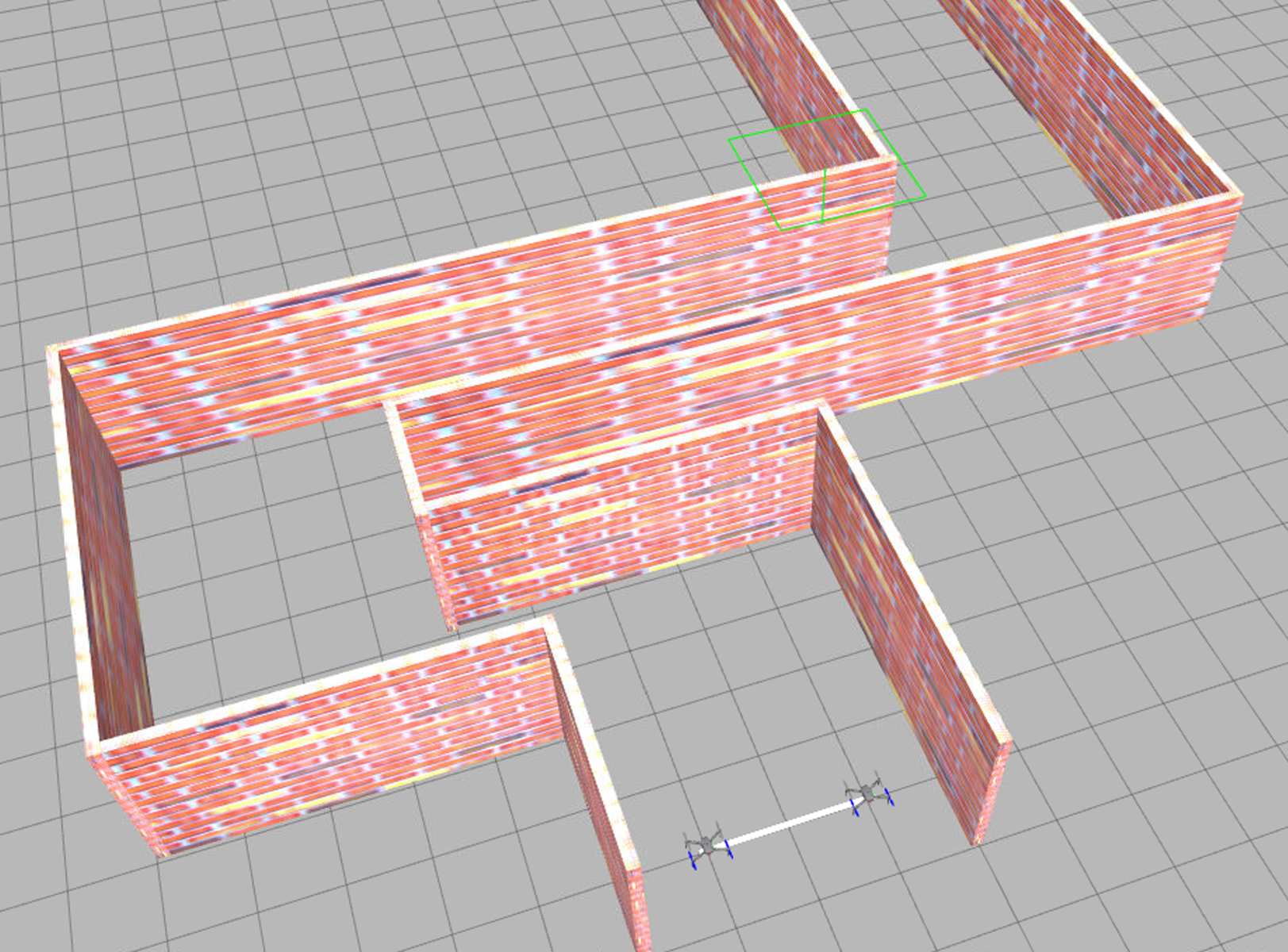}
				\par\end{centering}
		}\subfloat[]{
			\begin{centering}
				\includegraphics[width=0.45\linewidth]{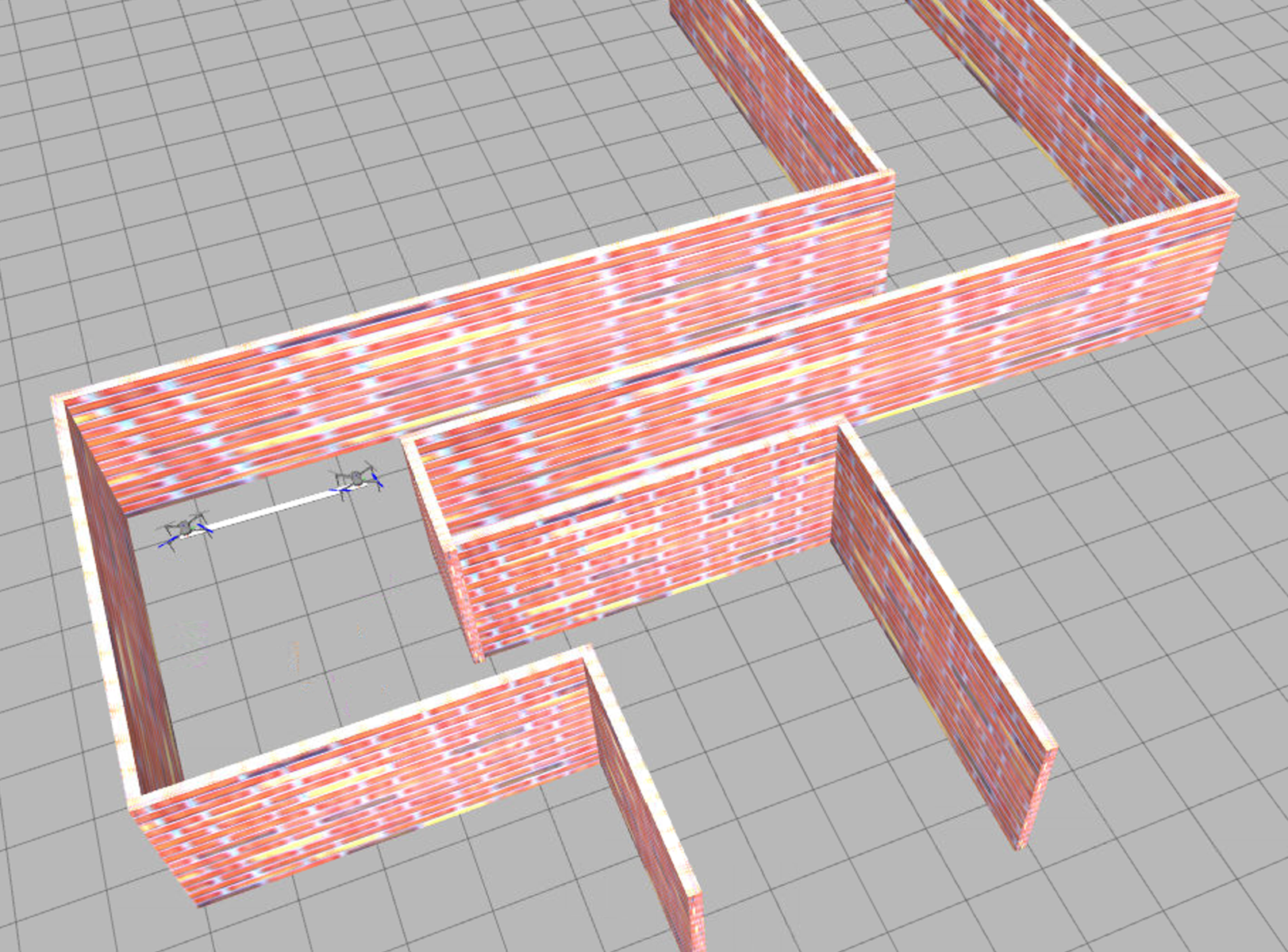}
				\par\end{centering}
			
		}
		\par\end{centering}
	\centering{}\subfloat[]{
		\begin{centering}
			\includegraphics[width=0.45\linewidth]{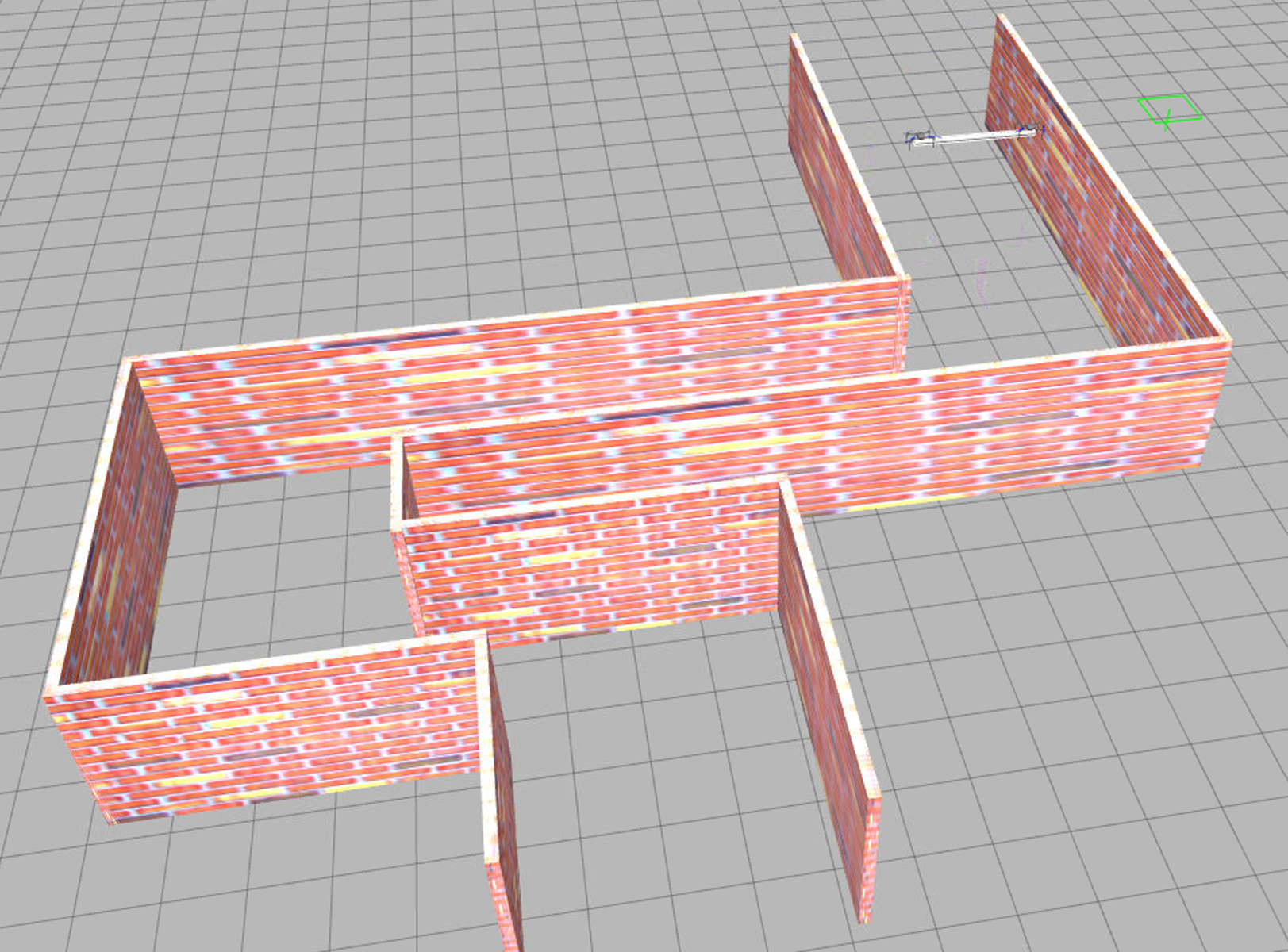}
			\par\end{centering}
		
	} \subfloat[]{\begin{centering}
			\includegraphics[width=0.45\linewidth]{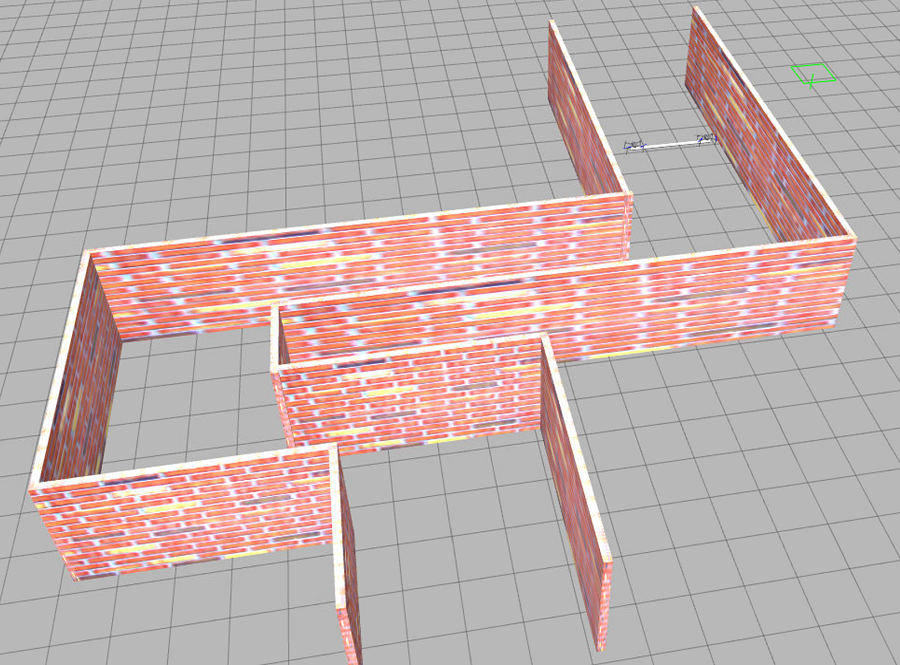}
			\par\end{centering}
		
	} \caption{\label{fig9}The results of simulating the system in ROS and Gazebo
		environment; (a) The starting position at the beginning of a zigzag
		corridor, (b) The system navigating through the zigzag corridor, (c)
		The system reaching the end of the zigzag corridor, (d) The system
		landing on the ground in the destination point at the end of a zigzag
		corridor.}
\end{figure*}

Finally, Figure \ref{fig9} shows the results of simulating the system
using ROS and Gazebo environments. Figure \ref{fig9}.(a) shows the
system at the starting position landing on the ground at the beginning
of a zigzag corridor. Then, Figure \ref{fig9}.(b) shows the system
navigating through the zigzag corridor according to the magnitudes
and directions of the applies forces. Then Figure \ref{fig9}.(c)
shows the system reaching the end of the zigzag corridor. Finally,
Figure \ref{fig9}.(d) shows the system landing on the ground at the
destination point at the end of a zigzag corridor. The human guidance
was simulated by applying forces in any direction through the Gazebo
interface. 

\section{Conclusion\label{Conclusion}}

In conclusion, this paper presents an assistive payload transportation
using a dual-quadrotor UAVs system with physical human interaction.
A novel Admittance-NFTSMC controller is utilized to control and stabilize
the system to track human guidance. The stability of the proposed
controller is proved using Lyapunov analysis. The proposed control
strategy was validated through different numerical simulations conducted
in MATLAB, ROS, and the Gazebo environments. The proposed system demonstrates
its ability to perform well in terms of stability, fast response,
and error convergence, highlighting its potential for practical implementation
in industrial applications with unstructured environments. Future
research directions include further optimization of the admittance
controller, integration of advanced sensing and perception capabilities,
and real-world experimentation to validate the effectiveness of the
proposed approach.

\section{Disclosure Statement}

No potential conflict of interest was reported by the author(s).

\section{Data Availability Statement}

The authors confirm that the data supporting the findings of this
study are available with the supplementary materials of the article.
The codes are available from the corresponding author upon reasonable
request.

\section{Acknowledgment}

This work was supported in part by the National Sciences and Engineering
Research Council of Canada (NSERC) under the grants RGPIN-2022-04937.
The authors also acknowledge the support of the University of Thi-Qar,
Iraqi Ministry of Higher Education and Scientific Research under financial
support (No. 6608 in 21/06/2022). 

\balance
\bibliographystyle{IEEEtran}
\bibliography{references}

\begin{thebibliography}{10}
\providecommand{\url}[1]{#1}
\csname url@samestyle\endcsname
\providecommand{\newblock}{\relax}
\providecommand{\bibinfo}[2]{#2}
\providecommand{\BIBentrySTDinterwordspacing}{\spaceskip=0pt\relax}
\providecommand{\BIBentryALTinterwordstretchfactor}{4}
\providecommand{\BIBentryALTinterwordspacing}{\spaceskip=\fontdimen2\font plus
\BIBentryALTinterwordstretchfactor\fontdimen3\font minus
  \fontdimen4\font\relax}
\providecommand{\BIBforeignlanguage}[2]{{%
\expandafter\ifx\csname l@#1\endcsname\relax
\typeout{** WARNING: IEEEtran.bst: No hyphenation pattern has been}%
\typeout{** loaded for the language `#1'. Using the pattern for}%
\typeout{** the default language instead.}%
\else
\language=\csname l@#1\endcsname
\fi
#2}}
\providecommand{\BIBdecl}{\relax}
\BIBdecl

\bibitem{ref1}
S.~Khalid, F.~Alnajjar, M.~Gochoo, A.~Renawi, and S.~Shimoda, ``Robotic
  assistive and rehabilitation devices leading to motor recovery in upper limb:
  a systematic review,'' \emph{Disability and Rehabilitation: Assistive
  Technology}, vol.~18, no.~5, pp. 658--672, 2023.

\bibitem{hasan2017cost}
H.~Hasan, ``A cost effective deaf-mute electronic assistant system using myo
  armband and smartphone,'' \emph{International Journal of Science and Research
  (IJSR)}, vol.~6, pp. 950--954, 2017.

\bibitem{naser2023internet}
H.~Naser, H.~Hammood, and A.~Q. Migot, ``Internet-based smartphone system for
  after-stroke hand rehabilitation,'' in \emph{2023 International Conference on
  Engineering, Science and Advanced Technology (ICESAT)}.\hskip 1em plus 0.5em
  minus 0.4em\relax IEEE, 2023, pp. 69--74.

\bibitem{ref2}
R.~Fuentes-Alvarez, J.~H. Hernandez, I.~Matehuala-Moran, M.~Alfaro-Ponce,
  R.~Lopez-Gutierrez, S.~Salazar, and R.~Lozano, ``Assistive robotic
  exoskeleton using recurrent neural networks for decision taking for the
  robust trajectory tracking,'' \emph{Expert Systems with Applications}, vol.
  193, p. 116482, 2022.

\bibitem{ref3}
H.~N. Hasan, ``A wearable rehabilitation system to assist partially hand
  paralyzed patients in repetitive exercises*,'' \emph{Journal of Physics:
  Conference Series}, vol. 1279, no.~1, p. 012040, jul 2019.

\bibitem{ref4}
J.~Li and M.~Ensafjoo, ``It’s not uav, it’s me: Demographic and self-other
  effects in public acceptance of a socially assistive aerial manipulation
  system for fatigue management,'' \emph{International Journal of Social
  Robotics}, vol.~16, no.~1, pp. 227--243, 2024.

\bibitem{ref5}
M.~Alkaddour, M.~A. Jaradat, S.~Tellab, N.~Sherif, M.~Alvi, L.~Romdhane, and
  K.~S. Hatamleh, ``Novel design of lightweight aerial manipulator for solar
  panel cleaning applications,'' \emph{IEEE Access}, 2023.

\bibitem{ref6}
Z.~Ghelichi, M.~Gentili, and P.~B. Mirchandani, ``Drone logistics for uncertain
  demand of disaster-impacted populations,'' \emph{Transportation research part
  C: emerging technologies}, vol. 141, p. 103735, 2022.

\bibitem{rodriguez2024inspection}
D.~A. Rodr{\'\i}guez, C.~L. Tafur, P.~F.~M. Daza, J.~A.~V. Vidales, and
  J.~C.~D. Rinc{\'o}n, ``Inspection of aircrafts and airports using uas: a
  review,'' \emph{Results in Engineering}, p. 102330, 2024.

\bibitem{ref9}
P.~Radoglou-Grammatikis, P.~Sarigiannidis, T.~Lagkas, and I.~Moscholios, ``A
  compilation of uav applications for precision agriculture,'' \emph{Computer
  Networks}, vol. 172, p. 107148, 2020.

\bibitem{meesaragandla2024herbicide}
S.~Meesaragandla, M.~P. Jagtap, N.~Khatri, H.~Madan, and A.~A. Vadduri,
  ``Herbicide spraying and weed identification using drone technology in modern
  farms: A comprehensive review,'' \emph{Results in Engineering}, p. 101870,
  2024.

\bibitem{ref10}
G.~Albeaino, M.~Gheisari, and R.~R. Issa, ``Human-drone interaction (hdi):
  Opportunities and considerations in construction,'' \emph{Automation and
  robotics in the architecture, engineering, and construction industry}, pp.
  111--142, 2022.

\bibitem{ref11}
J.~M. Nwaogu, Y.~Yang, A.~P. Chan, and H.-l. Chi, ``Application of drones in
  the architecture, engineering, and construction (aec) industry,''
  \emph{Automation in Construction}, vol. 150, p. 104827, 2023.

\bibitem{ranjbar2023addressing}
H.~Ranjbar, P.~Forsythe, A.~A.~F. Fini, M.~Maghrebi, and T.~S. Waller,
  ``Addressing practical challenge of using autopilot drone for asphalt surface
  monitoring: Road detection, segmentation, and following,'' \emph{Results in
  Engineering}, vol.~18, p. 101130, 2023.

\bibitem{ref12}
M.~A. Cheema, R.~I. Ansari, N.~Ashraf, S.~A. Hassan, H.~K. Qureshi, A.~K.
  Bashir, and C.~Politis, ``Blockchain-based secure delivery of medical
  supplies using drones,'' \emph{Computer Networks}, vol. 204, p. 108706, 2022.

\bibitem{ref13}
I.~H.~B. Pizetta, A.~S. Brand{\~a}o, and M.~Sarcinelli-Filho, ``Load
  transportation by quadrotors in crowded workspaces,'' \emph{IEEE Access},
  vol.~8, pp. 223\,941--223\,951, 2020.

\bibitem{ref15}
Z.~Ghelichi, M.~Gentili, and P.~B. Mirchandani, ``Logistics for a fleet of
  drones for medical item delivery: A case study for louisville, ky,''
  \emph{Computers \& Operations Research}, vol. 135, p. 105443, 2021.

\bibitem{hashim2023exponentially}
H.~A. Hashim, ``Exponentially stable observer-based controller for
  {VTOL}-{UAV}s without velocity measurements,'' \emph{International Journal of
  Control}, vol.~96, no.~8, pp. 1946--1960, 2023.

\bibitem{ref16}
D.~K. Villa, A.~S. Brandao, and M.~Sarcinelli-Filho, ``A survey on load
  transportation using multirotor uavs,'' \emph{Journal of Intelligent \&
  Robotic Systems}, vol.~98, pp. 267--296, 2020.

\bibitem{hashim2023observer}
H.~A. Hashim, A.~E. Eltoukhy, and A.~Odry, ``Observer-based controller for
  {VTOL}-{UAV}s tracking using direct vision-aided inertial navigation
  measurements,'' \emph{ISA transactions}, vol. 137, pp. 133--143, 2023.

\bibitem{ref20}
H.~Xie, K.~Dong, and P.~Chirarattananon, ``Cooperative transport of a suspended
  payload via two aerial robots with inertial sensing,'' \emph{IEEE Access},
  vol.~10, pp. 81\,764--81\,776, 2022.

\bibitem{ref21}
P.-X. Wu, C.-C. Yang, and T.-H. Cheng, ``Cooperative transportation of uavs
  without inter-uav communication,'' \emph{IEEE/ASME Transactions on
  Mechatronics}, 2023.

\bibitem{ref22}
X.~Liang, Z.~Su, W.~Zhou, G.~Meng, and L.~Zhu, ``Fault-tolerant control for the
  multi-quadrotors cooperative transportation under suspension failures,''
  \emph{Aerospace Science and Technology}, vol. 119, p. 107139, 2021.

\bibitem{shevidi2024quaternion}
A.~Shevidi and H.~A. Hashim, ``Quaternion-based adaptive backstepping fast
  terminal sliding mode control for quadrotor {UAV}s with finite time
  convergence,'' \emph{Results in Engineering}, p. 102497, 2024.

\bibitem{ref23}
Y.~Liu, F.~Zhang, P.~Huang, and X.~Zhang, ``Analysis, planning and control for
  cooperative transportation of tethered multi-rotor uavs,'' \emph{Aerospace
  Science and Technology}, vol. 113, p. 106673, 2021.

\bibitem{ref24}
B.~Shirani, M.~Najafi, and I.~Izadi, ``Cooperative load transportation using
  multiple uavs,'' \emph{Aerospace Science and Technology}, vol.~84, pp.
  158--169, 2019.

\bibitem{ref25}
S.~Barawkar, M.~Kumar, and M.~Bolender, ``Decentralized adaptive controller for
  multi-drone cooperative transport with offset and moving center of gravity,''
  \emph{Aerospace Science and Technology}, vol. 145, p. 108960, 2024.

\bibitem{ref26}
J.~Cai and B.~Xian, ``Robust hierarchical geometry control for the multiple
  uavs aerial transportation system with a suspended payload,'' \emph{Nonlinear
  Dynamics}, pp. 1--21, 2024.

\bibitem{hashim2023uwb}
H.~A. Hashim, A.~E. Eltoukhy, and K.~G. Vamvoudakis, ``{UWB} ranging and {IMU}
  data fusion: Overview and nonlinear stochastic filter for inertial
  navigation,'' \emph{IEEE Transactions on Intelligent Transportation Systems},
  2023.

\bibitem{fu2023iterative}
J.~Espin, C.~Camacho, and O.~Camacho, ``Control of non-self-regulating
  processes with long time delays using hybrid sliding mode control
  approaches,'' \emph{Results in Engineering}, vol.~22, p. 102113, 2024.

\bibitem{munoz2019adaptive}
R.~Alika, E.~M. Mellouli, and E.~H. Tissir, ``A modified sliding mode
  controller based on fuzzy logic to control the longitudinal dynamics of the
  autonomous vehicle,'' \emph{Results in Engineering}, p. 102120, 2024.

\bibitem{pisano2011sliding}
L.~Medina, G.~Guerra, M.~Herrera, L.~Guevara, and O.~Camacho, ``Trajectory
  tracking for non-holonomic mobile robots: A comparison of sliding mode
  control approaches,'' \emph{Results in Engineering}, p. 102105, 2024.

\bibitem{said2023application}
M.~said Adouairi, B.~Bossoufi, S.~Motahhir, and I.~Saady, ``Application of
  fuzzy sliding mode control on a single-stage grid-connected pv system based
  on the voltage-oriented control strategy,'' \emph{Results in Engineering},
  vol.~17, p. 100822, 2023.

\bibitem{ref17}
A.~Tagliabue, M.~Kamel, S.~Verling, R.~Siegwart, and J.~Nieto, ``Collaborative
  transportation using mavs via passive force control,'' in \emph{2017 IEEE
  International Conference on Robotics and Automation (ICRA)}, 2017, pp.
  5766--5773.

\bibitem{ref36}
J.~Horyna, T.~Baca, and M.~Saska, ``Autonomous collaborative transport of a
  beam-type payload by a pair of multi-rotor helicopters,'' in \emph{2021
  International Conference on Unmanned Aircraft Systems (ICUAS)}, 2021, pp.
  1139--1147.

\bibitem{ref19}
G.~Loianno and V.~Kumar, ``Cooperative transportation using small quadrotors
  using monocular vision and inertial sensing,'' \emph{IEEE Robotics and
  Automation Letters}, vol.~3, no.~2, pp. 680--687, 2018.

\bibitem{hashim2020nonlinear}
H.~A. Hashim and F.~L. Lewis, ``Nonlinear stochastic estimators on the special
  euclidean group {SE} (3) using uncertain imu and vision measurements,''
  \emph{IEEE Transactions on Systems, Man, and Cybernetics: Systems}, vol.~51,
  no.~12, pp. 7587--7600, 2022.

\bibitem{hashim2020systematic}
H.~A. Hashim, ``Systematic convergence of nonlinear stochastic estimators on
  the special orthogonal group {SO} (3),'' \emph{International Journal of
  Robust and Nonlinear Control}, vol.~30, no.~10, pp. 3848--3870, 2020.

\bibitem{ref37}
A.~Rajaeizadeh, A.~Naghash, and A.~Mohamadifard, ``Cooperative aerial payload
  transportation using two quadrotors,'' in \emph{Proceedings of the
  International micro air vehicle conference and flight competition}, 2017, pp.
  73--80.

\bibitem{ref38}
D.~Mellinger, M.~Shomin, N.~Michael, and V.~Kumar, ``Cooperative grasping and
  transport using multiple quadrotors,'' in \emph{Distributed Autonomous
  Robotic Systems: The 10th International Symposium}.\hskip 1em plus 0.5em
  minus 0.4em\relax Springer, 2013, pp. 545--558.

\bibitem{ref33}
F.~Augugliaro and R.~D'Andrea, ``Admittance control for physical
  human-quadrocopter interaction,'' in \emph{2013 European Control Conference
  (ECC)}.\hskip 1em plus 0.5em minus 0.4em\relax IEEE, 2013, pp. 1805--1810.

\bibitem{ref27}
P.~Prajapati and V.~Vashista, ``Aerial physical human robot interaction for
  payload transportation,'' \emph{IEEE Robotics and Automation Letters}, 2023.

\bibitem{ref29}
M.~Xu, A.~Hu, and H.~Wang, ``Visual-impedance-based human–robot
  cotransportation with a tethered aerial vehicle,'' \emph{IEEE Transactions on
  Industrial Informatics}, vol.~19, no.~10, pp. 10\,356--10\,365, 2023.

\bibitem{ref30}
M.~Romano, A.~Ye, J.~Pye, and E.~Atkins, ``Cooperative multilift slung load
  transportation using haptic admittance control guidance,'' \emph{Journal of
  Guidance, Control, and Dynamics}, vol.~45, no.~10, pp. 1899--1912, 2022.

\bibitem{hashim2019special}
H.~A. Hashim, ``Special {O}rthogonal {G}roup {SO}(3), {E}uler {A}ngles,
  {A}ngle-axis, {R}odriguez {V}ector and {U}nit-quaternion: {O}verview,
  {M}apping and {C}hallenges,'' \emph{arXiv preprint arXiv:1909.06669}, 2019.

\bibitem{hashim2019nItoStrat}
H.~A. Hashim, L.~J. Brown, and K.~McIsaac, ``Nonlinear stochastic attitude
  filters on the special orthogonal group 3: {I}to and {S}tratonovich,''
  \emph{IEEE Transactions on Systems, Man, and Cybernetics: Systems}, vol.~49,
  no.~9, pp. 1853--1865, 2019.

\bibitem{ref28}
M.~Boukattaya, N.~Mezghani, and T.~Damak, ``Adaptive nonsingular fast terminal
  sliding-mode control for the tracking problem of uncertain dynamical
  systems,'' \emph{ISA transactions}, vol.~77, pp. 1--19, 2018.

\bibitem{ref32}
M.~Tognon, R.~Alami, and B.~Siciliano, ``Physical human-robot interaction with
  a tethered aerial vehicle: Application to a force-based human guiding
  problem,'' \emph{IEEE Transactions on Robotics}, vol.~37, no.~3, pp.
  723--734, 2021.

\end{thebibliography}

\end{document}